%% file: iclr2021_conference.tex
\documentclass{article} 
\usepackage{iclr2021_conference,times}

\input{math_commands.tex}

\usepackage{hyperref}
\usepackage{url}
\usepackage{amsmath}
\usepackage{amsfonts}
\usepackage{subcaption}
\usepackage{xcolor}
\usepackage{graphicx}
\usepackage{algorithm}
\usepackage{algorithmicx}
\usepackage{algpseudocode}
\usepackage[switch]{lineno}
\usepackage{booktabs}
\usepackage{wrapfig}
\usepackage{listings}
\newcommand{\ourM}{M-CURL}
\title{Masked Contrastive Representation Learning for Reinforcement Learning}

\author{
Jinhua Zhu$^{1}$, Yingce Xia$^{2}$, Lijun Wu$^2$,\\
\textbf{\,Jiajun Deng$^{1}$,Wengang Zhou$^{1}$, Tao Qin$^{2}$, Houqiang Li$^{1}$} \\
$^1$University of Science and Technology of China;\\ $^2$Microsoft Research; \\
$^1$\texttt{\{teslazhu,dengjj\}@mail.ustc.edu.cn},\;\texttt{\{zhwg,lihq\}@ustc.edu.cn}\\
$^2$\texttt{\{yingce.xia,lijuwu,taoqin\}@microsoft.com}}

\iclrfinalcopy 
\begin{document}

\maketitle

\begin{abstract}
Improving sample efficiency is a key research problem in reinforcement learning (RL), and CURL, which uses contrastive learning to extract high-level features from raw pixels of individual video frames, is an efficient algorithm~\citep{srinivas2020curl}. We observe that consecutive video frames in a game are highly correlated but CURL deals with them independently. To further improve data efficiency, we propose a new algorithm, masked contrastive representation learning for RL, that takes the correlation among consecutive inputs into consideration. In addition to the CNN encoder and the policy network in CURL, 
our method introduces an auxiliary Transformer module to leverage the correlations among video frames. During training, we randomly mask the features of several frames, and use the CNN encoder and Transformer to reconstruct them based on the context frames. The CNN encoder and Transformer are jointly trained via contrastive learning where the reconstructed features should be similar to the ground-truth ones while dissimilar to others. 
During inference, the CNN encoder and the policy network are used to take actions, and the Transformer module is discarded. Our method achieves consistent improvements over CURL on $14$ out of $16$ environments from DMControl suite and $21$ out of $26$ environments from Atari 2600 Games. The code is available at \url{https://github.com/teslacool/m-curl}.
\end{abstract}
\section{Introduction}
Recently, reinforcement learning (RL) has achieved great success in game AI~\citep{silver2018general,vinyals2019grandmaster}, robotics~\citep{gu2017deep,bousmalis2018using}, logistic~\citep{li2019cooperative}, etc. In many settings, the dimensions of the raw inputs are high, which is hard to directly work on and sample-inefficient~\citep{oord2018representation,srinivas2020curl}. Alternatively, we can first learn an encoder, to map the high-dimensional input into low-dimensional representations, which will be further fed into the policy networks to get the actions. How to efficiently learn the encoder is an important problem for RL, and numerous methods have been proposed to address this problem. Among them, one line is introducing an auxiliary loss for learning better latent representations~\citep{jaderberg2016reinforcement,oord2018representation,lee2019stochastic,yarats2019improving}, and another line is using one world model to sample rollouts and plan in the learned pixel or latent space~\citep{hafner2019learning,hafner2019dream,kaiser2019model}.

Recently, contrastive learning has attracted more and more attention in machine learning and computer vision communities~\citep{he2020momentum,henaff2019data}. Contrastive learning is an instance-level pre-training technique to get an encoder. Taking image processing as an example, we first use the encoder to extract representations for a collection of images, which are saved into a dictionary. An encoded query should be similar to the matching one in the dictionary and dissimilar to others. Such an idea is adapted into the first line aforementioned~\citep{oord2018representation,srinivas2020curl} due to its efficiency. One of the representative methods is CURL~\citep{srinivas2020curl}, which is short for contrastive unsupervised representations for RL. CURL aims to learn the encoder in a self-supervised manner, where an input $x$ is augmented (e.g., random crop) into two variants $x_q$ (query) and $x_+$ (matching key). The negative samples $x_{-}$'s are randomly selected from the replay buffer~\citep{mnih2013playing}, where $x$ is excluded. After processed by the encoder, $x_q$ should be similar to $x_+$ and dissimilar to the sampled $x_{-}$'s.

Although CURL makes great success in pixel-based RL, it overlooks the relevance among the inputs. After receiving an input from the environment, we take an action, get an instant reward, and observe another input. However, in pixel-based RL games, consecutive frames are highly correlated while this property is ignored. Intuitively, a good encoder should not only provide an accurate representation of the current input, but also remember the previous inputs and be predictable for future ones.

In this paper, we propose a new algorithm, masked contrastive representation learning for RL (\ourM), to further improve data efficiency and performance of pixel-based RL. Apart from the convolutional neural network (CNN) as image encoder used in previous works (e.g., CURL), we introduce an auxiliary Transformer module~\citep{vaswani2017attention}, the state-of-the-art structure to model sequences, to capture the relevance. To achieve that, we first sample a set of consecutive transitions from the replay buffer and regard them as a sequence. We then use the CNN encoder to map each selected frames (i.e., images) into low-dimensional representations. Inspired by the success of masked pre-training in natural language processing like BERT~\citep{devlin2018bert} and RoBERTa~\citep{lample2019cross}, we randomly mask out several representations and let Transformer module reconstruct them. Finally, we use contrastive loss to jointly train the CNN encoder and Transformer, where the reconstructed features should be similar to the ground-truth ones while dissimilar to others.

The contribution of our work is summarized as follows: (1) We propose \ourM, an effective representation learning method for RL leveraging the advantages of masked pre-training and contrastive learning; (2) Our method is a general one and can be combined with any off-policy algorithm with a replay buffer, such as SAC \citep{haarnoja2018soft} and rainbown DQN \citep{mnih2013playing} (we can also add one replay buffer for on-policy algorithms, such as A3C \citep{mnih2016asynchronous}). (3) \ourM{} surpasses CURL by a large margin on $14$ out of $16$ environments from DMControl suite and $21$ out of $26$ environments from Atari 2600 Games, which is the previously state-of-the-art algorithm leveraging instance discrimination only.

The remaining parts are organized as follows: Several related work and background are introduced in Section~\ref{sec:related_work}. We introduce our method in detail in Section~\ref{sec:our_method}, and present some experiment results and analyses in Section~\ref{sec:experiments} and Section~\ref{sec:ablation}. Last, we conclude our work and propose some future works in Section~\ref{sec:conclusion}.

\section{Background and Related work}\label{sec:related_work}
In this section, we introduce the background and related work of contrastive learning and sample-efficient RL.

\noindent \textbf{Contrastive Learning}: Contrastive learning is a self-supervised algorithm for representation learning, which does not require additional labeling on a dataset. Contrastive learning can be regarded as an instance classification problem, where one instance should be distinguished from others. Taking image processing as an example, denote $\mathcal{I}=\{I_1,I_2,\cdots,I_N\}$ as a collection of $N$ images. $f_q$ and $f_k$ are CNN encoders, both of which can map images into hidden representations. To use contrastive learning, we first use $f_k$ to build a look-up dictionary $\mathcal{K}=\{ f_k(I_i)|I_i\in\mathcal{I}\}$. Then, given an query image $I_i, i\in[N]$, we use the query encoder $f_q$ to encode $I_i$ and get $q_i=f_q(I_i)$. $q_i$ should be similar to $k_i$ (i.e., the matching key in the dictionary) while dissimilar to those in $\mathcal{K}\backslash\{k_i\}$. \citet{oord2018representation} proposed InfoNCE, which is an effective contrastive loss function. Mathematically,
\begin{equation}
    \mathcal{L}_{\text{infoNCE}}(q_i,\mathcal{K})=-\log \frac{\exp \left(q_i \cdot k_i / \tau\right)} {\sum_{k \in \mathcal{K} }\exp \left(q_i \cdot k/ \tau\right)},
\end{equation}
where $\tau$ is a hyperparamter. $f_q$ and $f_k$ can share parameters~\citep{wu2018unsupervised} or not~\citep{he2020momentum}. Contrastive learning has achieved great success in image processing~\citep{wu2018unsupervised,henaff2019data,he2020momentum,tian2019contrastive}, and has been extended to graph neural networks~\citep{sun2019infograph,hassani2020contrastive}, speech translation \citep{khurana2020cstnet}, etc. Specifically, \citet{srinivas2020curl} introduced contrastive loss into image based RL problems to enhance the image encoder, which has proven to be helpful for data-efficiency on DMControl and Atari benchmark.

\noindent \textbf{Sample-efficient RL} Sample-efficiency is one of the long-standing challenges in real-world applications like robotics and control. While learning a policy network from high-dimensional inputs with limited samples is much challenging, some works have taken meaningful steps \citep{hafner2019learning,hafner2019dream,srinivas2020curl,laskin2020reinforcement,kostrikov2020image}. To effectively leverage the data, introducing auxiliary tasks to enhance the representations is an active topic in RL.  \citet{shelhamer2016loss,yarats2019improving} use generative modeling to reconstruct the original input in pixel space according to the hidden representations. \citet{shelhamer2016loss,oord2018representation,lee2019stochastic} propose to predict the future frames  based on the representations of past observations and actions. CURL~\citep{srinivas2020curl} first propose to leverage contrastive learning on individual frames to get one better encoder. Recently, some works introduce data augmentation into RL, and the policy should be invariant to different views~\citep{laskin2020reinforcement,kostrikov2020image}. Our work is complementary to them and we will make a combination in the future. Another line in sample-efficiency RL is to sample rollouts and plan \citep{hafner2019learning,hafner2019dream,kaiser2019model} according to a learnable world model. Later, we will discuss the relation of our method with these works.

\section{Our Method}\label{sec:our_method}
In this work, we focus on image based RL problems. We briefly review the preliminaries of this problem in Section~\ref{sec:prelim}, then introduce our method in detail in Section~\ref{sec:alg} and provide some discussions in Section~\ref{sec:discussion}.

\subsection{Preliminaries}\label{sec:prelim}
The image based RL tasks can be formulated as a partially observed Markov decision process~\citep{kostrikov2020image}, which is described by the tuple $(\mathcal{O}, \mathcal{A}, p, r, \gamma)$, with each element defined as follows: (1) $\mathcal{O}$ represents observations, a collection of images rendered from the environment. (2) $\mathcal{A}$ is the action space. Let $o_t \in \mathcal{O}$ and $a_t\in\mathcal{A}$ denote the rendered image and action at time step $t$. (3) $p$ denotes the transition dynamics, which represents the probability distribution over the next observation $o_t^\prime$ given $o_{\leq t}$ and $a_t$. Mathematically, $p=\operatorname{Pr}\left(o_{t}^{\prime} \mid o_{\leq t}, a_{t}\right)$. (4) $r: \mathcal{O} \times \mathcal{A} \mapsto \mathbb{R}$ is the reward function that maps the observation $o_{\leq t}$ and action $a_t$ to a reward value, i.e., $r_{t}=r\left(o_{\leq t}, a_{t}\right) \in \mathbb{R}$. (5) $\gamma\in(0,1]$ is the discount factor to trade off immediate and future rewards.

\begin{wrapfigure}{r}{0.4\textwidth}
\centering
\includegraphics[width=0.95\linewidth]{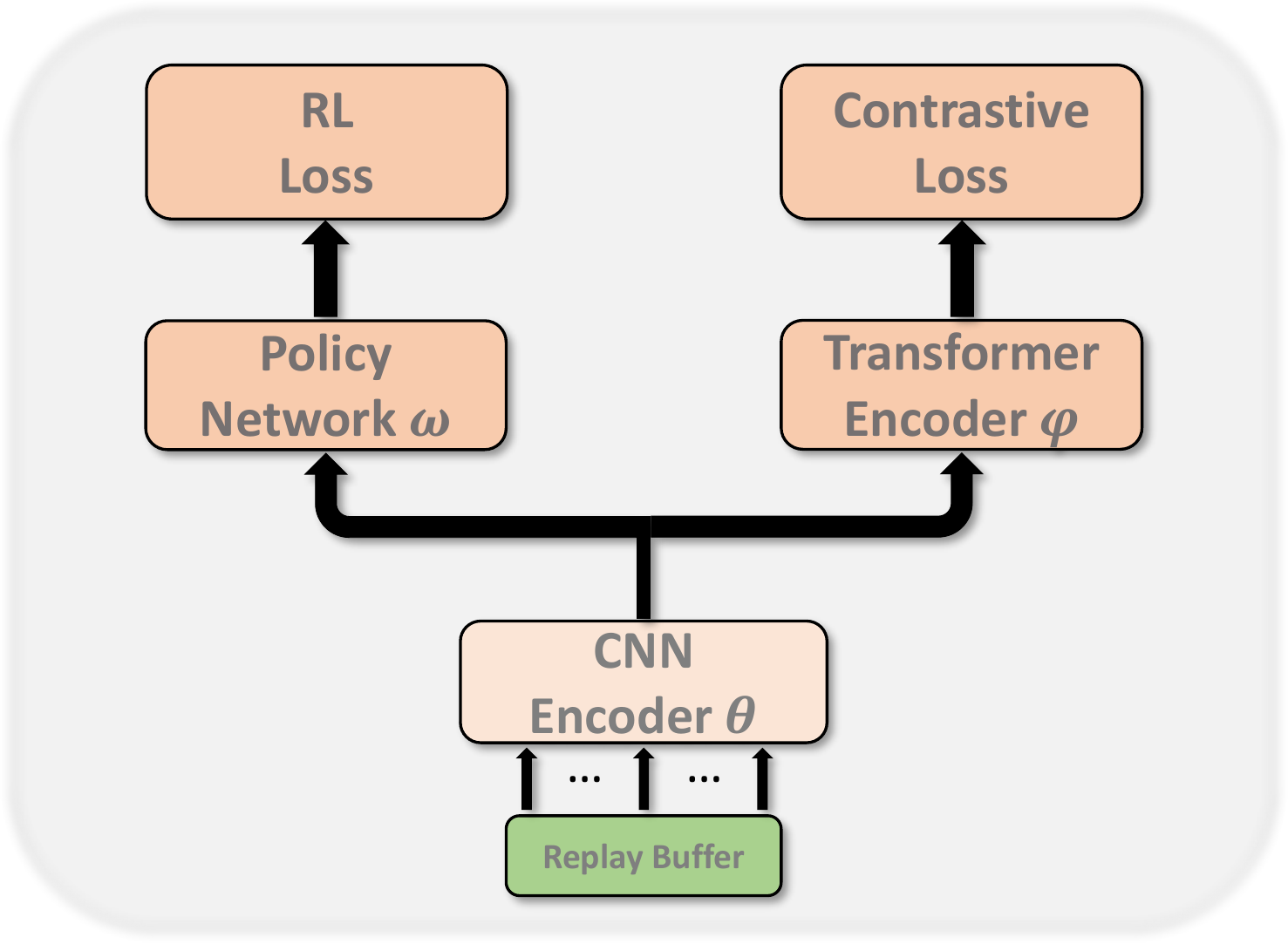}
\caption{The training framework of \ourM. The left part is the vanilla RL algorithm, such as SAC or DQN. The right part is a Transformer module to reconstruct the masked input.}
\label{fig:overall}
\end{wrapfigure}
In deep RL, we use neural networks to process the observations and make decisions. Following common practice~\citep{mnih2013playing}, we stack $K(>1)$ consecutive observations as the input to capture more information. That is, $s_t=(o_{t-K+1},o_{t-k+2},\cdots,o_t)$, and let $\mathcal{S}$ denote the collection of all $s_t$'s. Let $f_\theta$ denote the CNN encoder parameterized by $\theta$, which maps $s\in\mathcal{S}$ into a $d$-dimension representation. After that, the encoded representation will be fed into a policy network $\pi_\omega$ parameterized by $\omega$ to get the action at time step $t$. That is, $a_t=\pi(f_\theta(s_t))$, $a_t\in\mathcal{A}$.  The objective is to maximize the cumulative discounted return, which is defined as $\mathbb{E}_{\pi}\sum_{t=1}^{\infty} \gamma^{t} r_{t}$. The $\pi_\omega$ can be specialized as any model-free algorithm, like Soft Actor Critic (SAC) \citep{haarnoja2018soft}, Rainbow DQN~\citep{mnih2013playing},  etc.

\noindent{\it Replay buffer $\mathcal{B}$}: To stabilize the training, \citet{mnih2013playing} introduce a replay buffer $\mathcal{B}$ to deep Q-learning algorithms. At time step $t$, after observing $s_t$, we take action $a_t$, get reward $r_t$ and get a new observation $s_{t+1}$. We append the tuple $(s_t, a_t, s_{t+1}, r_t, d_{t})$ to $\mathcal{B}$, where $d_t$ indicates whether the episode terminates. The buffer has a maximum length, after which we will remove the oldest tuple from $\mathcal{B}$. In \citet{mnih2013playing}, the parameters of the Q-value networks are updated by minimizing $\sum_{j=1}^{B}(r_j + \gamma\max_{a'}Q(s_{j+1},a';\theta_Q^{'})-Q(s_j,a_j;\theta_Q))^2$, where $\theta_Q$ is the learnable parameter, and we need to randomly sample $B$ trajectories from $\mathcal{B}$. In CURL, the negative samples used to learn the CNN encoder are also randomly selected from $\mathcal{B}$.

\subsection{Algorithm}\label{sec:alg}

Inspired by the pre-training in natural language processing~\citep{devlin2018bert,lample2019cross}, we leverage the masked pre-training technique and the Transformer architecture here. The overall framework is shown in Figure~\ref{fig:overall}. A stack of $K$ images $s_t\in\mathcal{S}$ is first processed by $f_\theta:\mathcal{S}\mapsto\mathbb{R}^d$ to get a $d$-dimension representation. After that, there are two parts: the left part is a policy network $\pi_\omega:\mathbb{R} ^d\mapsto\mathcal{A}$, which outputs an action according to the current representation. The right part is an auxiliary model $\varphi$, which is a Transformer module and will be optimized via contrastive loss. $f_\theta$ and $\pi_\omega$ are introduced in  Section~\ref{sec:prelim}. Next, we will introduce the auxiliary model $\varphi$ in detail, including data preparation, model architecture and training objective function.

\noindent(1) {\bf Data preparation}: We sample $T$ consecutive observations from the replay buffer by chronological order. Denote the sampled observations as $S=(s_1,s_2,\cdots,s_T)$. To leverage the masked training technique, define $M=(M_1,M_2,\cdots,M_T)$, where for each $i\in[M]$, w.p. $\varrho_m$, $M_i=1$; w.p. $1-\varrho_m$, $M_i=0$. $\varrho_m\in[0,1]$ is a hyperparameter. If $M_i=1$, following~\citet{devlin2018bert}, $s_i$ is modified as follows: w.p. $80\%$, the $s_i$ is replaced with a zero vector; w.p. $10\%$, it is replaced by another $s_j$ randomly sampled from $\mathcal{B}$; w.p. $10\%$, it remains unchanged. The modified observation is denoted as $s_i^\prime$. Let $S^\prime=(\bar{s}_1^\prime,\bar{s}_2^\prime,\cdots,\bar{s}_T^\prime)$ denote the refined $S$, where $\bar{s}^\prime_i=s_i^\prime M_i + s_i(1-M_i)$. The $S^\prime$ is encoded by $f_\theta$ and we  get $H^0=(h_1,h_2,\cdots,h_T)$ where $h_i=f_\theta(\bar{s}_i^\prime)$, $\bar{s}_i^\prime\in S^\prime$. 

\noindent(2) {\bf Architectures of $\varphi$}: $\varphi$ is used to reconstruct the masked input by leveraging the global input. The architecture of the Transformer encoder is leveraged.  Specifically, $\varphi$ consists of $L$ identical blocks. Each block is stacked by two layers, including a self-attention layer $\operatorname{attn}(\cdots)$ and a feed-forward layer $\operatorname{ffn}(\cdots)$. Let $h^l_i$ denote the output of $i$-th position of the $l$-th block, and $H^l=(h^l_1,h^l_2,\cdots,h^l_T)$. For any $l\in[L]$, we have that
\begin{equation}
\begin{aligned}
\tilde{h}^l_i&=\operatorname{attn}(h^{l-1}_i,H^{l-1})=\sum_{i=1}^{T} \alpha_{i} W_{v} h^{l-1}_i,\\ 
\alpha_{i}&=\exp \left(\left(W_{q} q\right)^{T}\left(W_{k} k_{i}\right)\right)/Z,\\
Z&=\sum_{i=1}^{T} \exp \left(\left(W_{q} q\right)^{T}\left(W_{k} k_{i}\right)\right).
\end{aligned}
\label{eq:self_attn}
\end{equation}
Then $\tilde{h}^l_i$ is transformed via $\operatorname{ffn}$ and eventually we get
\begin{equation}
h_i^l=\operatorname{ffn}(\tilde{h}^l_i)=W_{2} \max \left(W_{1} \tilde{h}^l_i+b_{1}, 0\right)+b_{2}.
\label{eq:ffn}
\end{equation}
In \eqref{eq:self_attn} and \eqref{eq:ffn}, the $W$'s and $b$'s are the parameters to be learned. By enumerating $l$ from $1$ to $L$ and alternatively using \eqref{eq:self_attn} and \eqref{eq:ffn}, we will eventually obtain $H^L$. Note that positional embedding, layer normalization \citep{ba2016layer} and residual connections are all used. We leave the details in Appendix \ref{transformer_arch}. 

\begin{figure*}[!t]
\centering
\includegraphics[width=0.8\linewidth]{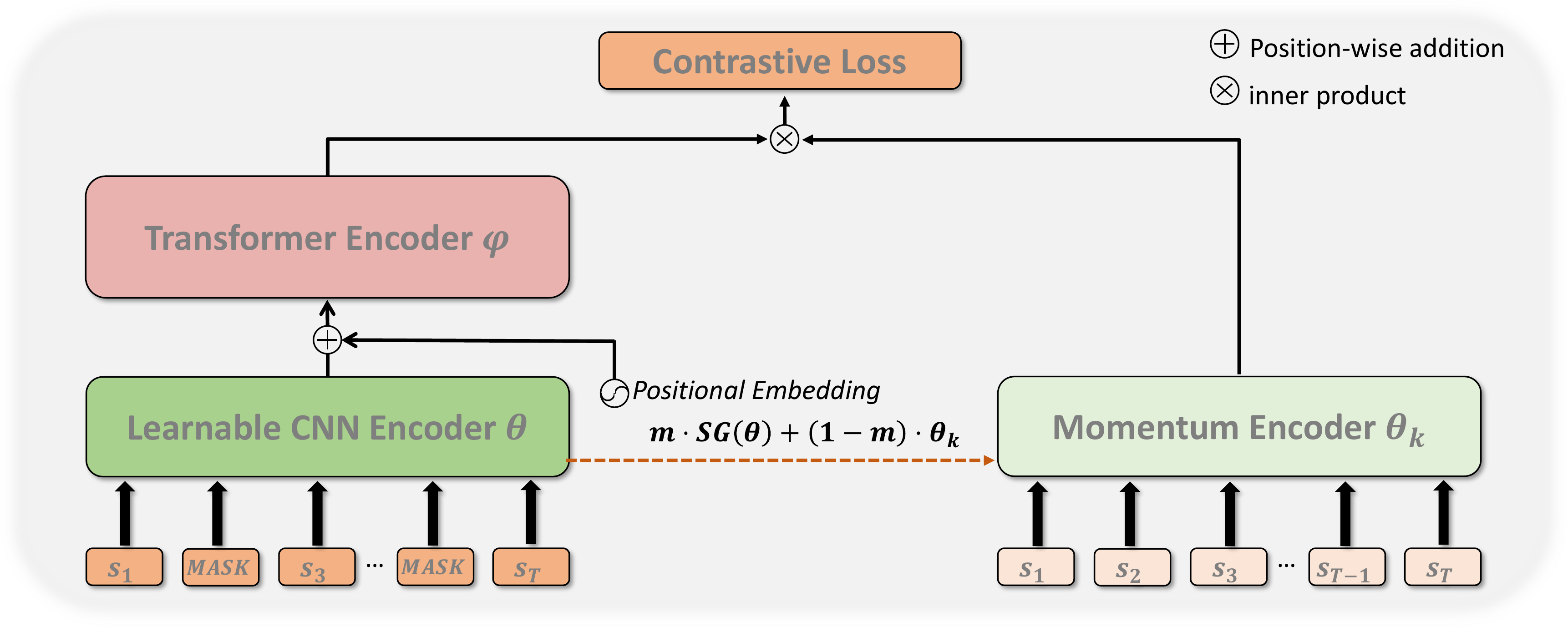}
\caption{Training strategy of \ourM.}
\vspace{-0.4cm}
\label{fig:ctmr}
\end{figure*}

\noindent(3) {\bf Training strategy}: Following~\citet{srinivas2020curl}, we use momentum contrastive learning to optimize $\varphi$ and $f_\theta$. The workflow of our training strategy is in Figure~\ref{fig:ctmr}. The query set $Q$ is the $H^L$ obtained in the previous step. The key set $K=(k_1,k_2,\cdots,K_T)$ is encoded from the non-masked set $S$: $k_i=f_{\theta_k}(s_i)$, $s_i\in S$. The architectures of query encoder $f_\theta$ and key encoder $f_{\theta_k}$ are the same but the parameters are different, where $\theta_k$ is iteratively updated as $\theta_k=m\texttt{SG}(\theta)+(1-m)\theta_k$, $m$ is the momentum value and $\texttt{SG}$ means ``stop gradient'' following~\citet{he2020momentum}.

The contrastive loss is defined as follows:
\begin{equation}
\mathcal{L}_{\text{ct}}= \sum_{i=1}^T -M_i \log \frac{\exp \left(q_i \cdot k_i/ \tau\right)} {\sum_{j=1}^T \exp \left(q_i \cdot k_j/ \tau\right)},
\label{eq:contrastive_loss_}
\end{equation}
where $\tau$ is a hyperparameter. The intuition behind \eqref{eq:contrastive_loss_} is that for any masked position, the reconstructed $h_i^L$ (i.e., $q_i$) should be similar to its original feature $k_i$, and dissimilar to the others. In this way, the $H^0$ output by $f_\theta$ should be relevant such that when we mask specific features in $H^0$, we can still reconstruct them by using the context features. 

For the RL training objective function, we can use any one for off-policy RL algorithms. Generally, the loss for RL in our paper can be summarized as
\begin{equation}
\begin{aligned}
\mathcal{L}_{\text{rl}} = &\mathbb{E}_{\left(s_t, a_t\right) \sim \mathcal{B}}\left[\left(Q_{\omega_1}\left(f_\theta(s_t), a_t\right)-\hat{Q}\left(f_{\theta^ \prime} (s_t), a_t\right)\right)^2 \right.\\
&\left. + \alpha\mathrm{D}_{\mathrm{KL}}\left(\pi_{\omega_2}\left(\cdot \mid f_\theta(s_t)\right) \| \frac{\exp \left(Q_{\omega_1}\left(f_\theta(s_t), \cdot\right)\right)}{Z_{\theta}\left(s_t\right)}\right)\right].
\end{aligned}
\label{eq:rl_obj}
\end{equation}
where $(s_t, a_t)$ is randomly sampled from $\mathcal{B}$, $\omega_1$ and $\omega_2$ are the parameters of the Q-function and the policy network, $\hat{Q}$ is the target value, $\mathrm{D}_{\mathrm{KL}}(\cdot\|\cdot)$ is 
KL-divergence between the policy function and Q-function \citep{haarnoja2018soft} and $Z_{\theta}\left(s_t\right)$ is the normalization term.

The overall objective function of our approach is
\begin{equation}
\min_{\theta,\omega,\varphi}\mathcal{L}_{\text{all}} = \mathcal{L}_{\text{rl}} + \lambda\mathcal{L}_{\text{ct}},
\label{eq:overall_obj}
\end{equation}
where $\theta$ is related to both terms in \eqref{eq:overall_obj}. Following~\citet{srinivas2020curl}, we fix $\lambda$ as $1$.
A detailed example of \ourM{}  coupling with SAC algorithm is described in Appendix \ref{algo_sac}. 

\subsection{Discussion}\label{sec:discussion}
Methods to improve sample efficiency in pixel-based RL can be divided into two categories, one is to sample rollouts and plan through a world model \citep{hafner2019learning,hafner2019dream,kaiser2019model}, and the other is to introduce auxiliary losses as discussed in previous section.
While showing improved sample efficiency, the former class usually have difficulty on balancing various auxiliary loss, such as the dynamic loss, reconstruction loss,  KL regularization term,  reward prediction loss in addition to value function and policy function optimization, and are correspondingly brittle to hyper-parameters tuning and hard to reproduce. In contrast, in our method, we only introduce an additional loss and it is still widely effective to fix the balance coefficient as $1$ across all our experiments. Our method belongs to the latter class, and has many advancements on it. First, different from reconstruction-based methods \citep{shelhamer2016loss,yarats2019improving}, which always optimize on a generative object on single observation, our method try to use one discriminative loss. Second, another related work is to use latent variable models to model the temporal structure in the MDPs (POMDPs) and get a compact and disentangled representations \citep{oord2018representation,lee2019stochastic}. However, these methods are all use the LSTMs \citep{hochreiter1997long} as their basic arhcitecture which limits their method to be unidirectional. In our method, we use Transformer to model the context relation, and to the best of our knowledge, our method is the first one to incorporate Transformer into RL to model this bidirectional characteristic.




\begin{figure}[!ht]
    \centering
    
    \begin{subfigure}[b]{0.245\textwidth}
    \centering
    \includegraphics[width=\textwidth]{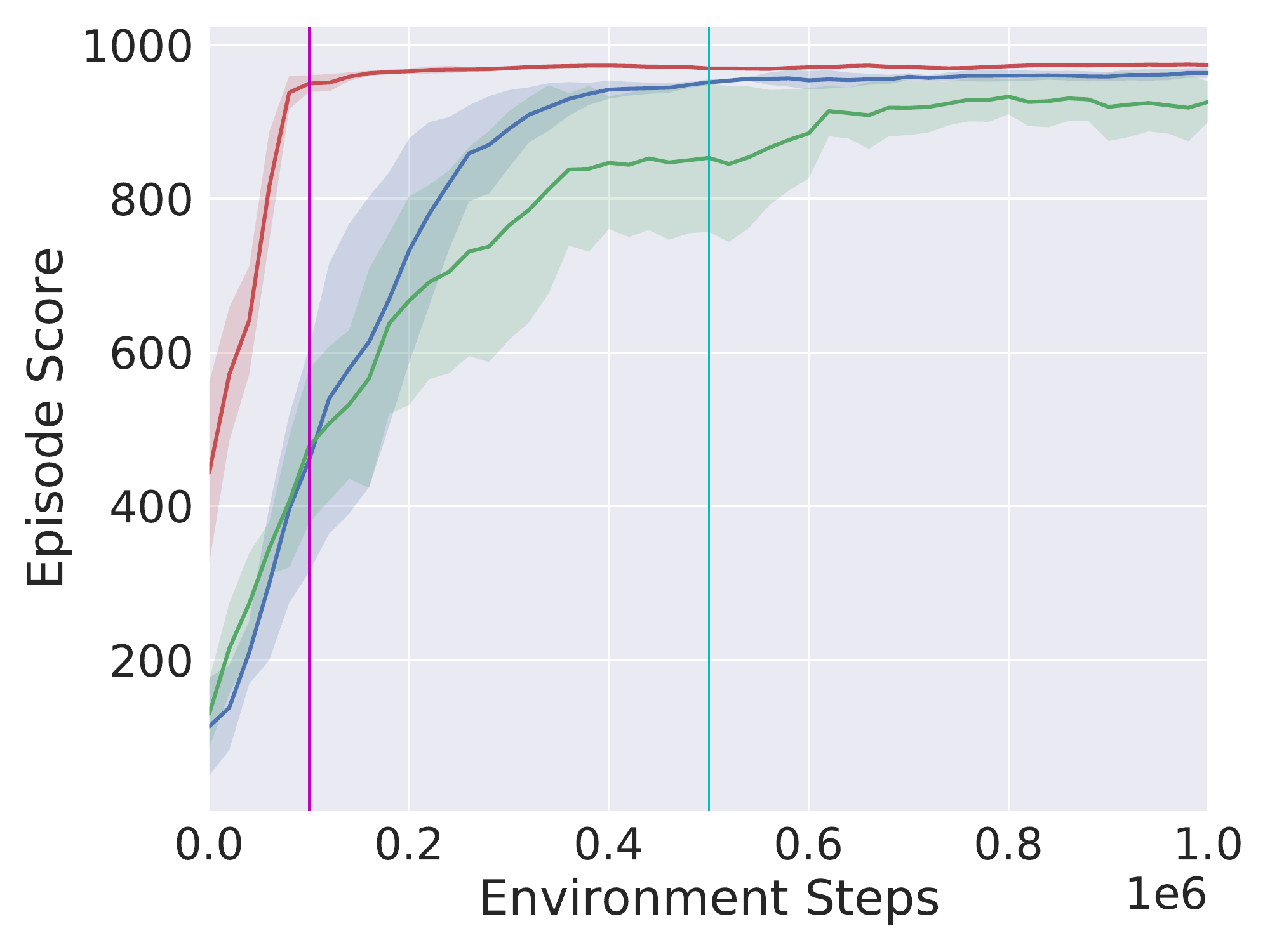}
    \caption{\small ball in cup catch}\label{dmbicc}
    \end{subfigure}
    \hfill
    \begin{subfigure}[b]{0.245\textwidth}
    \centering
    \includegraphics[width=\textwidth]{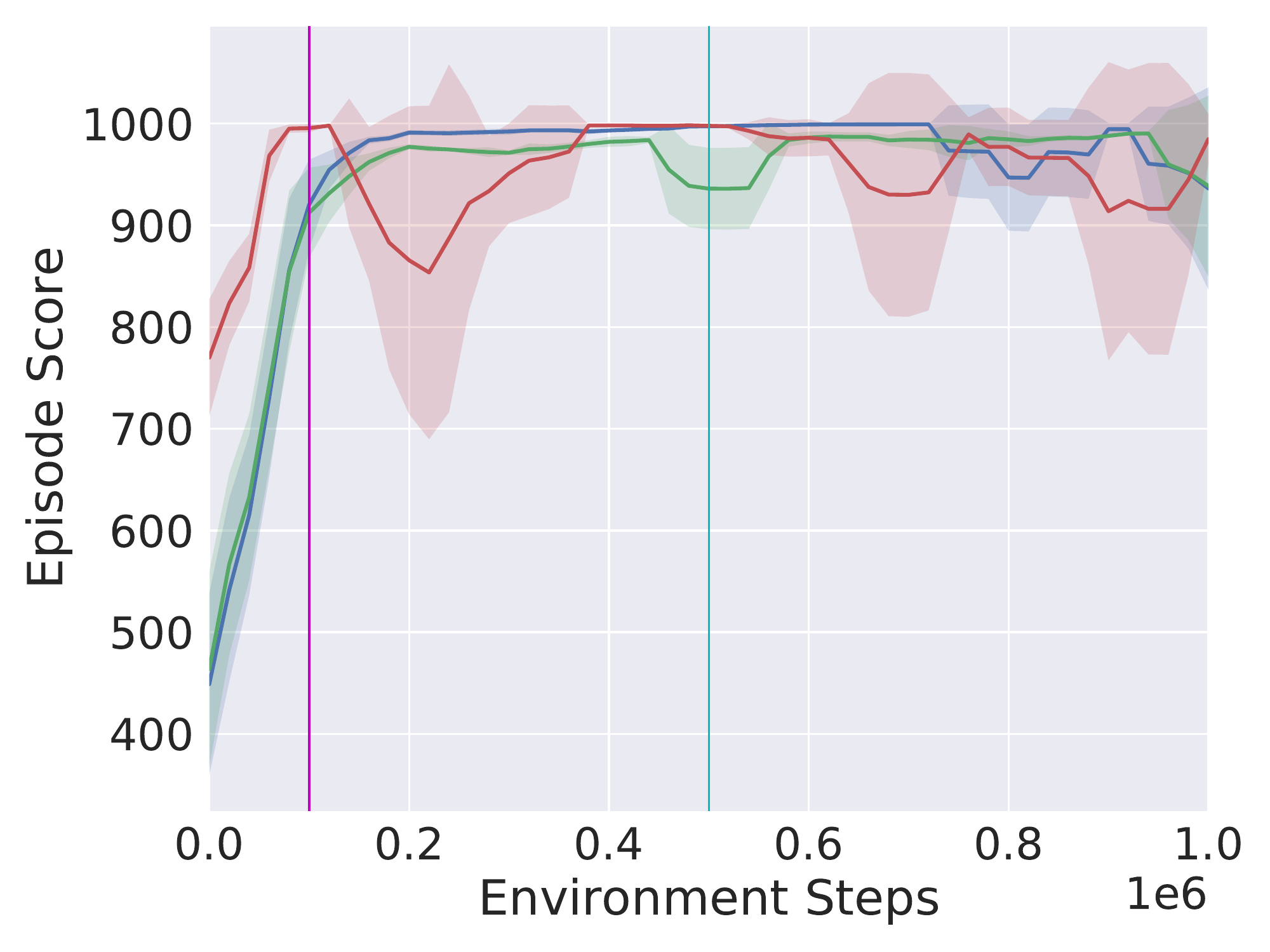}
    \caption{\small cartpole-balance}
    \end{subfigure}
    \hfill
    \begin{subfigure}[b]{0.245\textwidth}
    \centering
    \includegraphics[width=\textwidth]{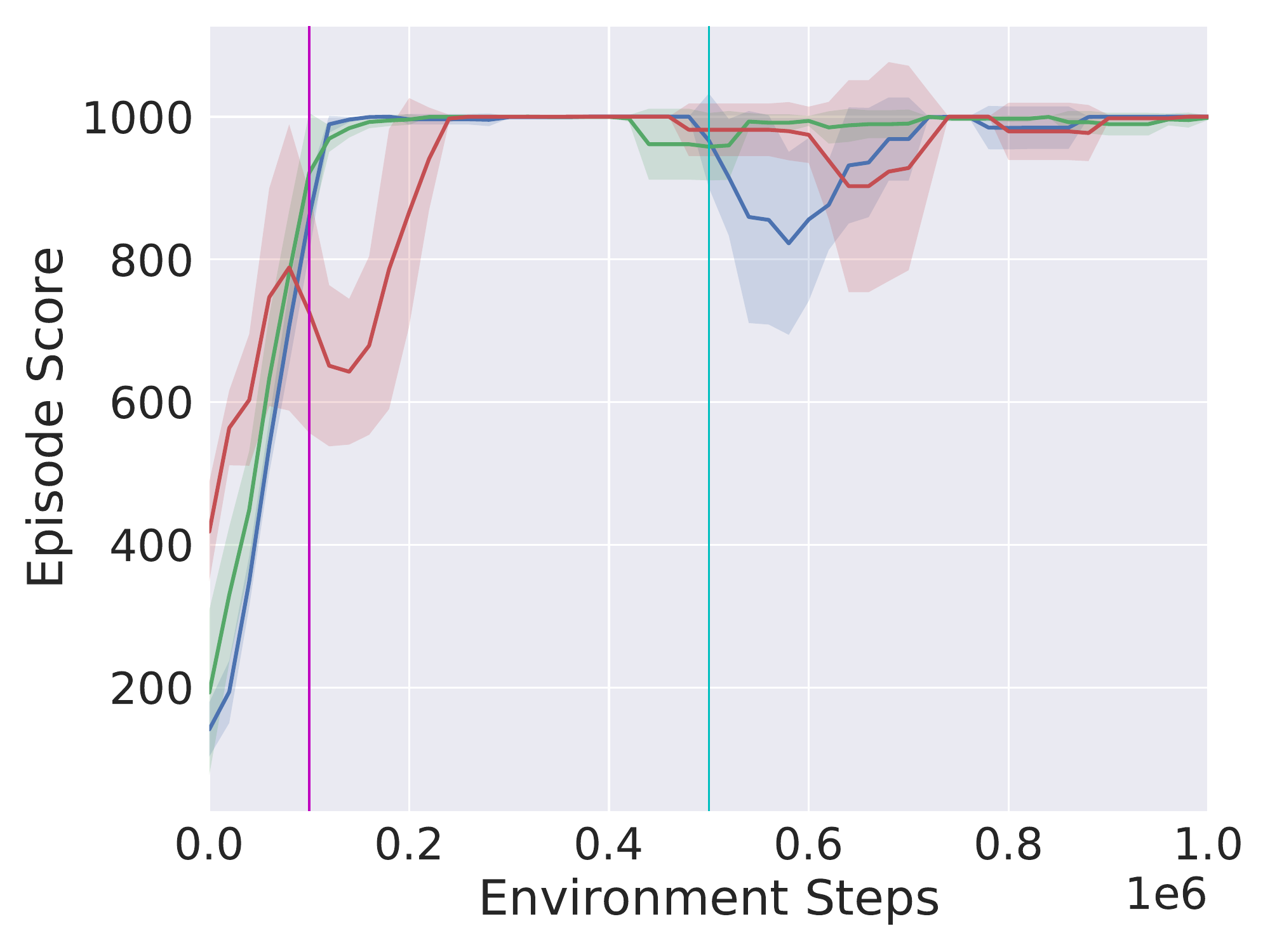}
    \caption{\small cartpole-balance sparse}
    \end{subfigure}
    \hfill
    \begin{subfigure}[b]{0.245\textwidth}
    \centering
    \includegraphics[width=\textwidth]{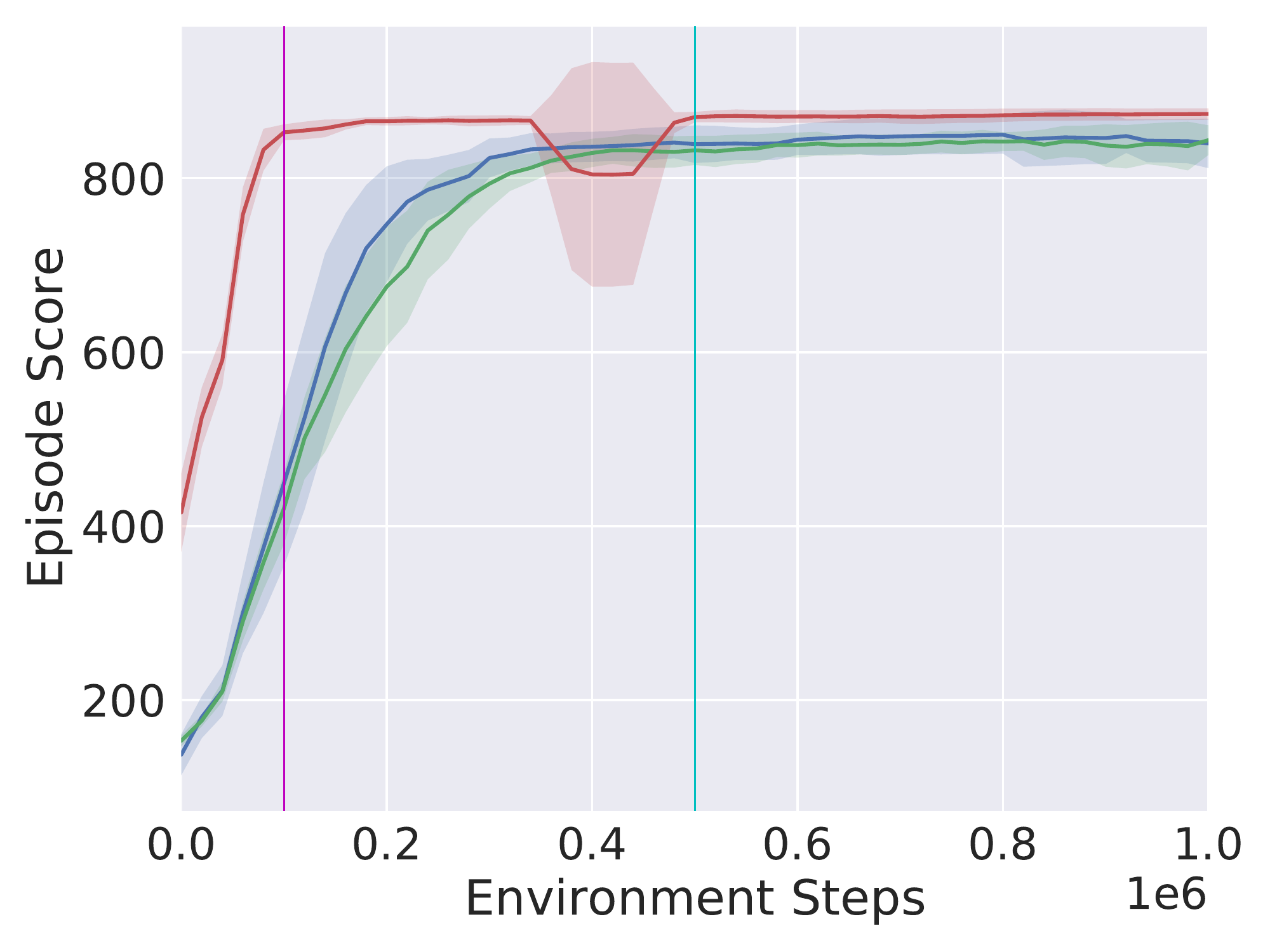}
    \caption{\small cartpole-swingup}
    \end{subfigure}
    
    \begin{subfigure}[b]{0.245\textwidth}
    \centering
    \includegraphics[width=\textwidth]{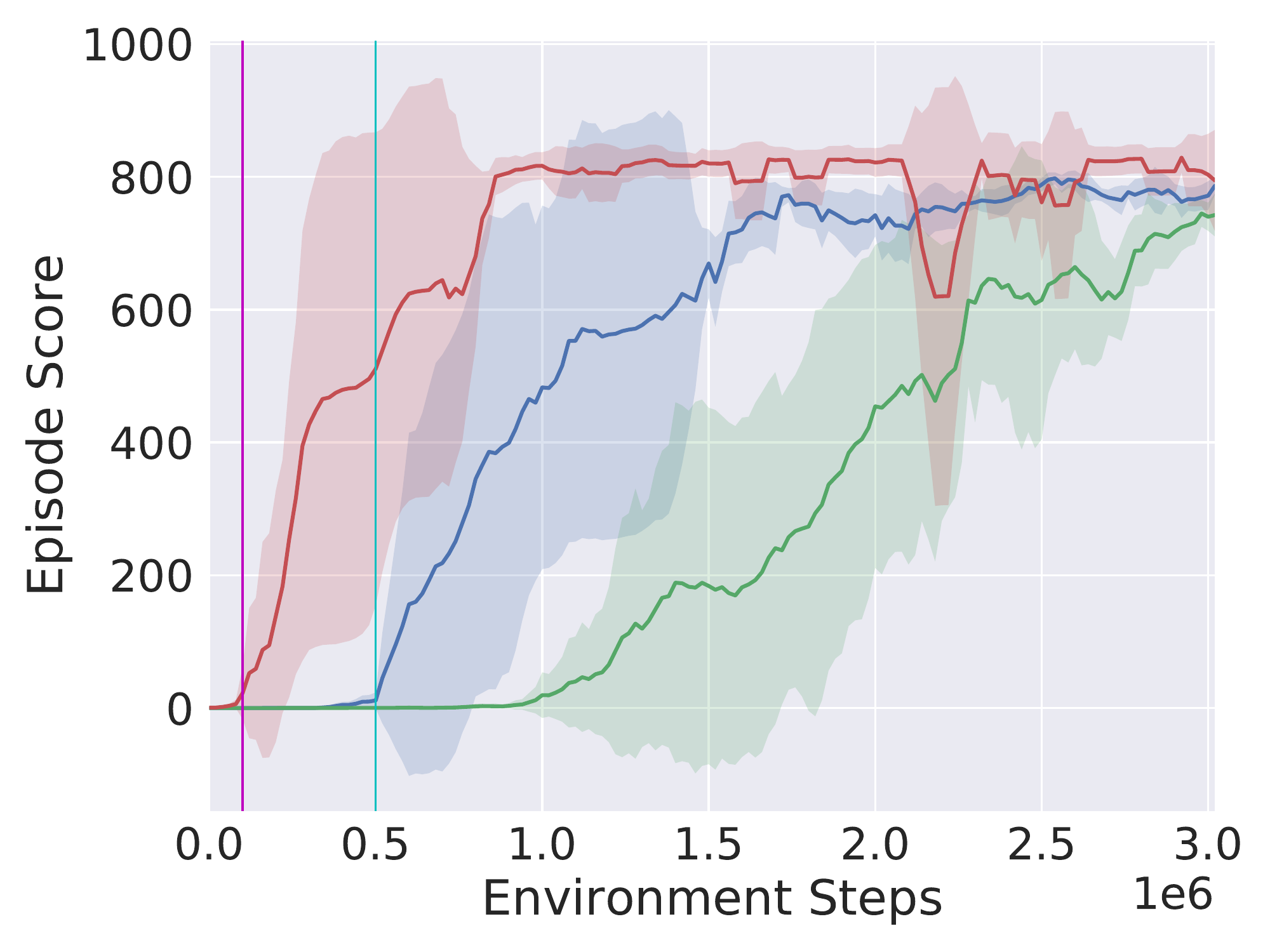}
    \caption{\small cartpole-swingup sparse}
    \end{subfigure}
    \hfill
    \begin{subfigure}[b]{0.245\textwidth}
    \centering
    \includegraphics[width=\textwidth]{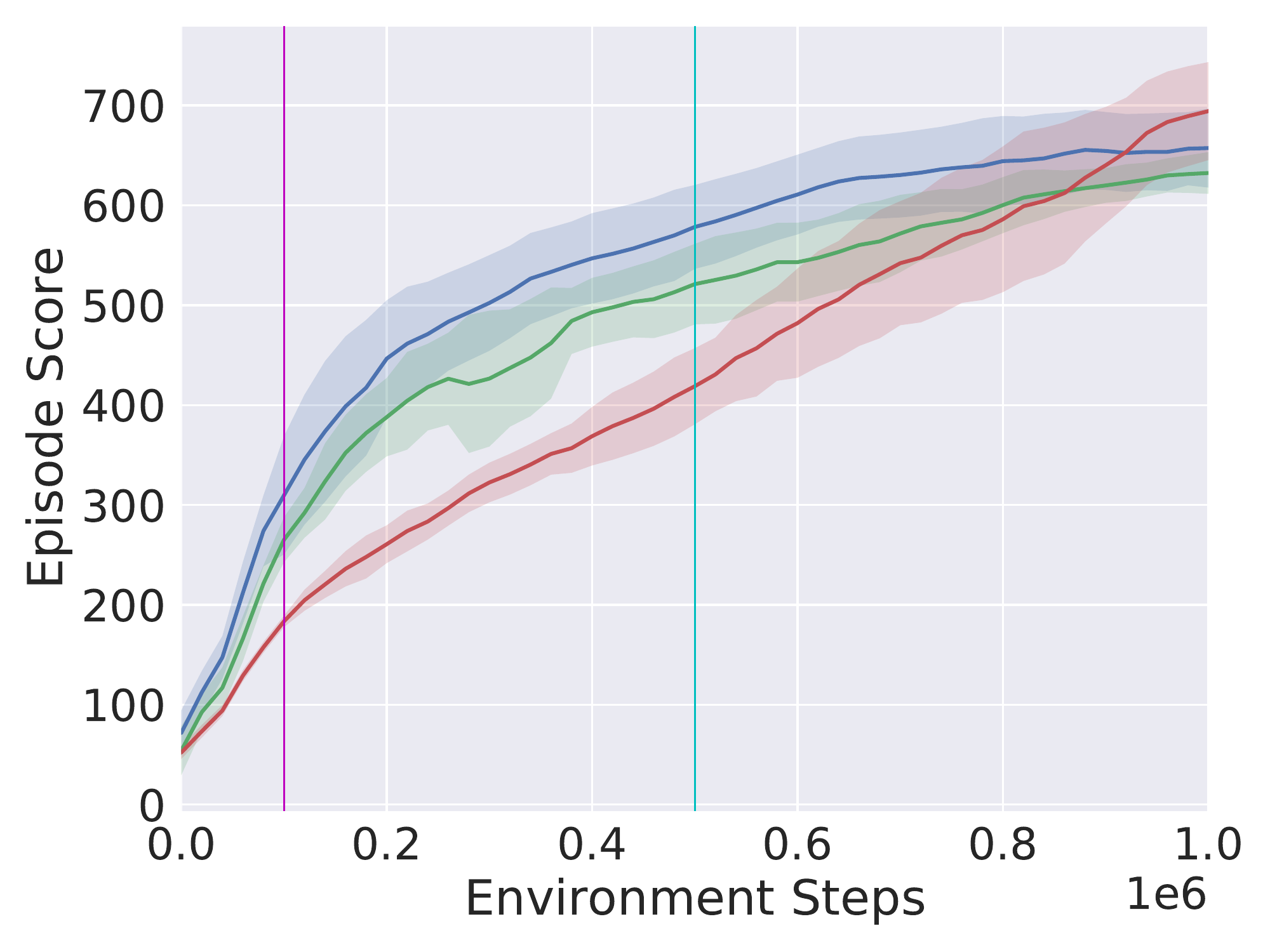}
    \caption{\small cheetah-run}
    \end{subfigure}
    \hfill
    \begin{subfigure}[b]{0.245\textwidth}
    \centering
    \includegraphics[width=\textwidth]{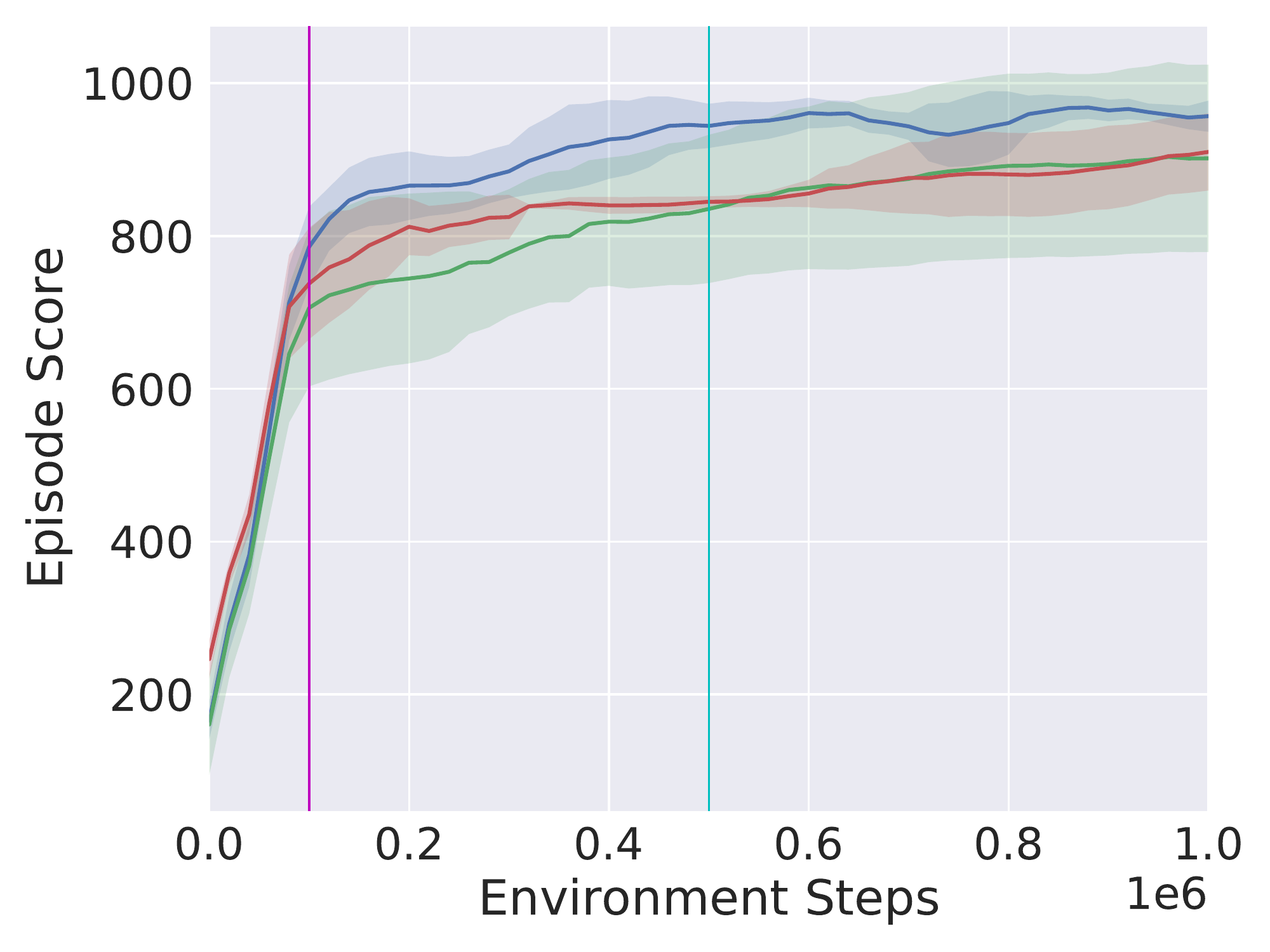}
    \caption{\small finger-spin}
    \end{subfigure}
    \hfill
    \begin{subfigure}[b]{0.245\textwidth}
    \centering
    \includegraphics[width=\textwidth]{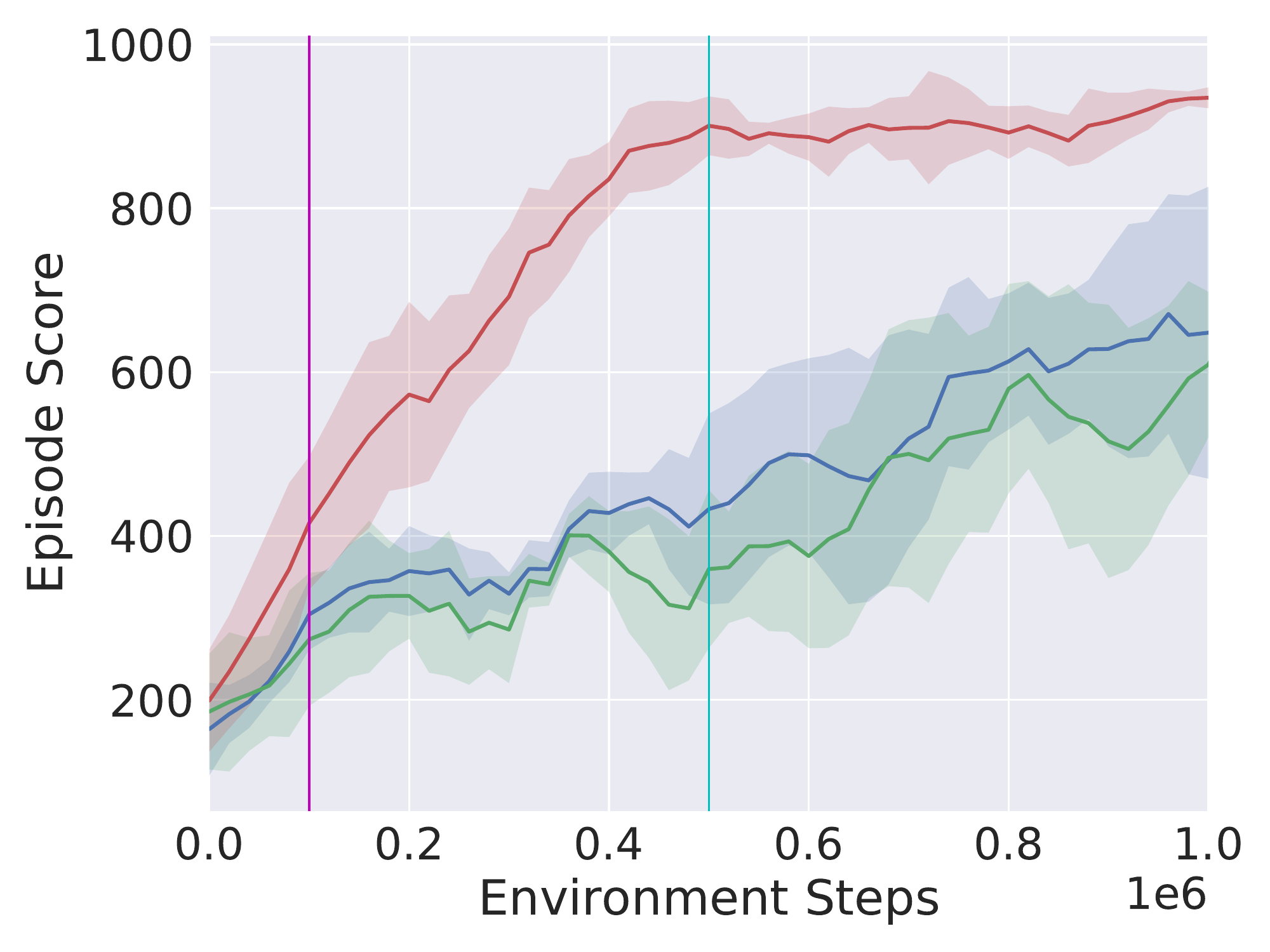}
    \caption{\small finger-turn easy}
    \end{subfigure}

    \begin{subfigure}[b]{0.245\textwidth}
    \centering
    \includegraphics[width=\textwidth]{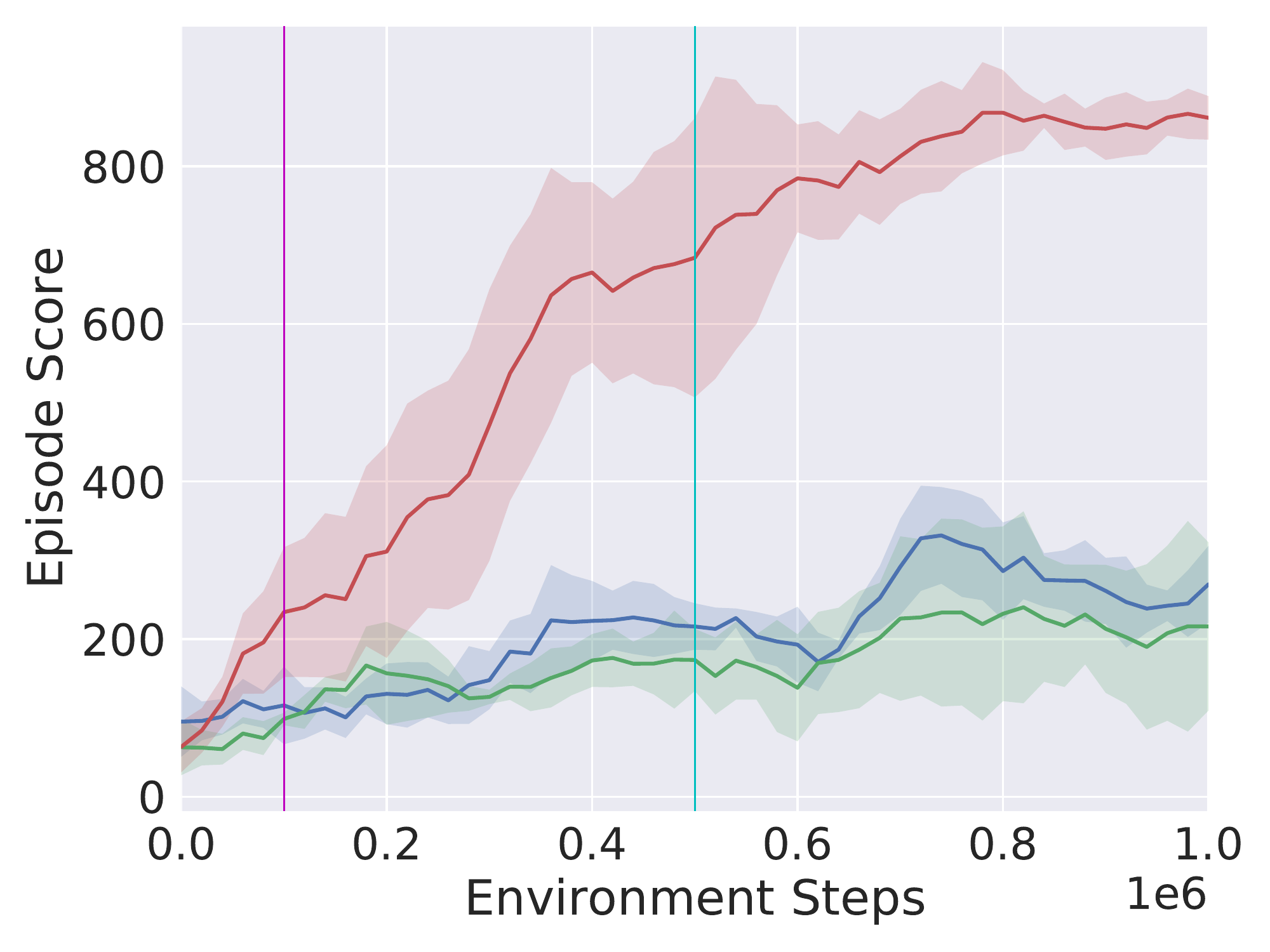}
    \caption{\small finger-turn hard}
    \end{subfigure}
    \hfill
    \begin{subfigure}[b]{0.245\textwidth}
    \centering
    \includegraphics[width=\textwidth]{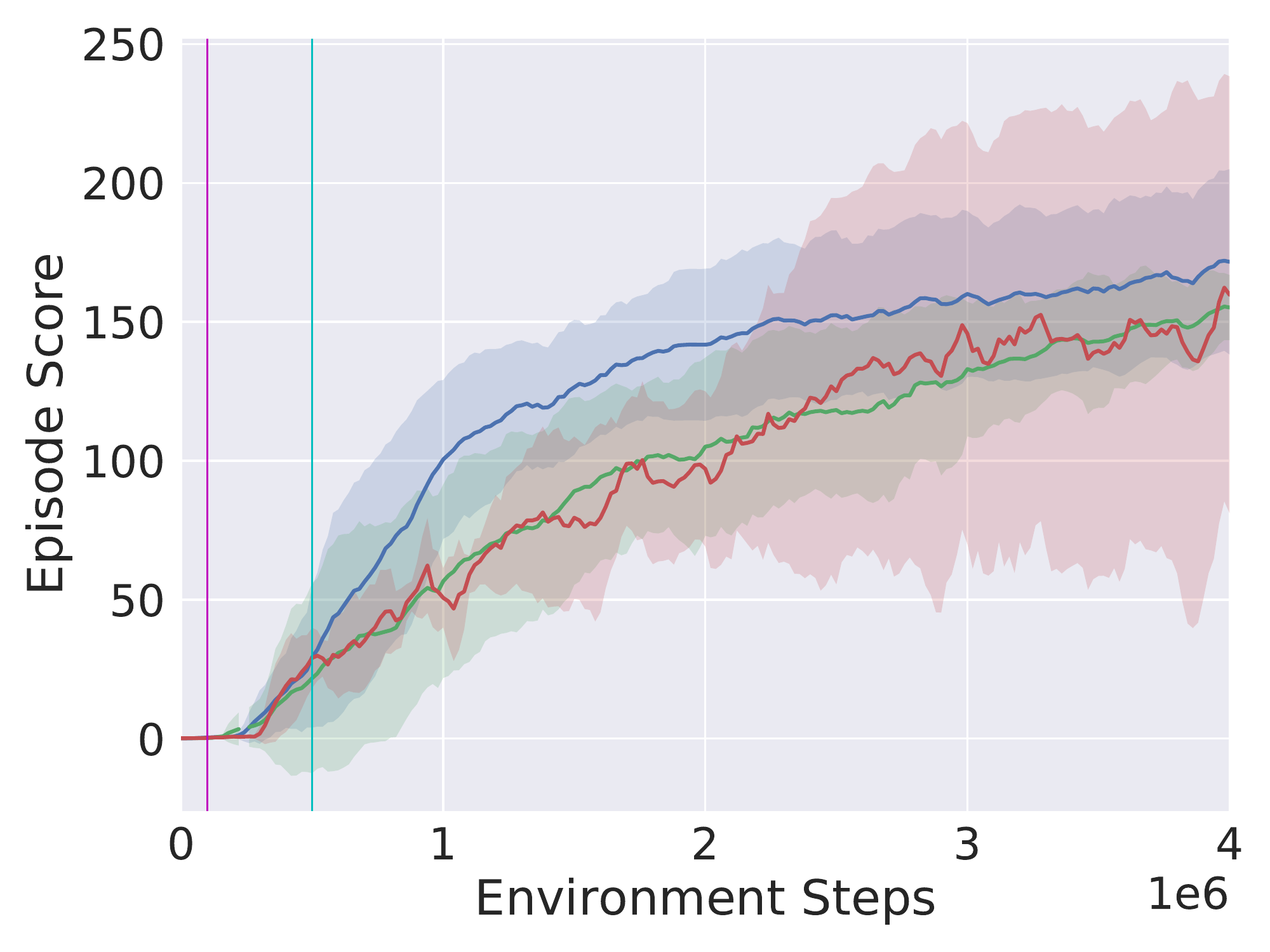}
    \caption{\small hopper-hop}
    \end{subfigure}
    \hfill
    \begin{subfigure}[b]{0.245\textwidth}
    \centering
    \includegraphics[width=\textwidth]{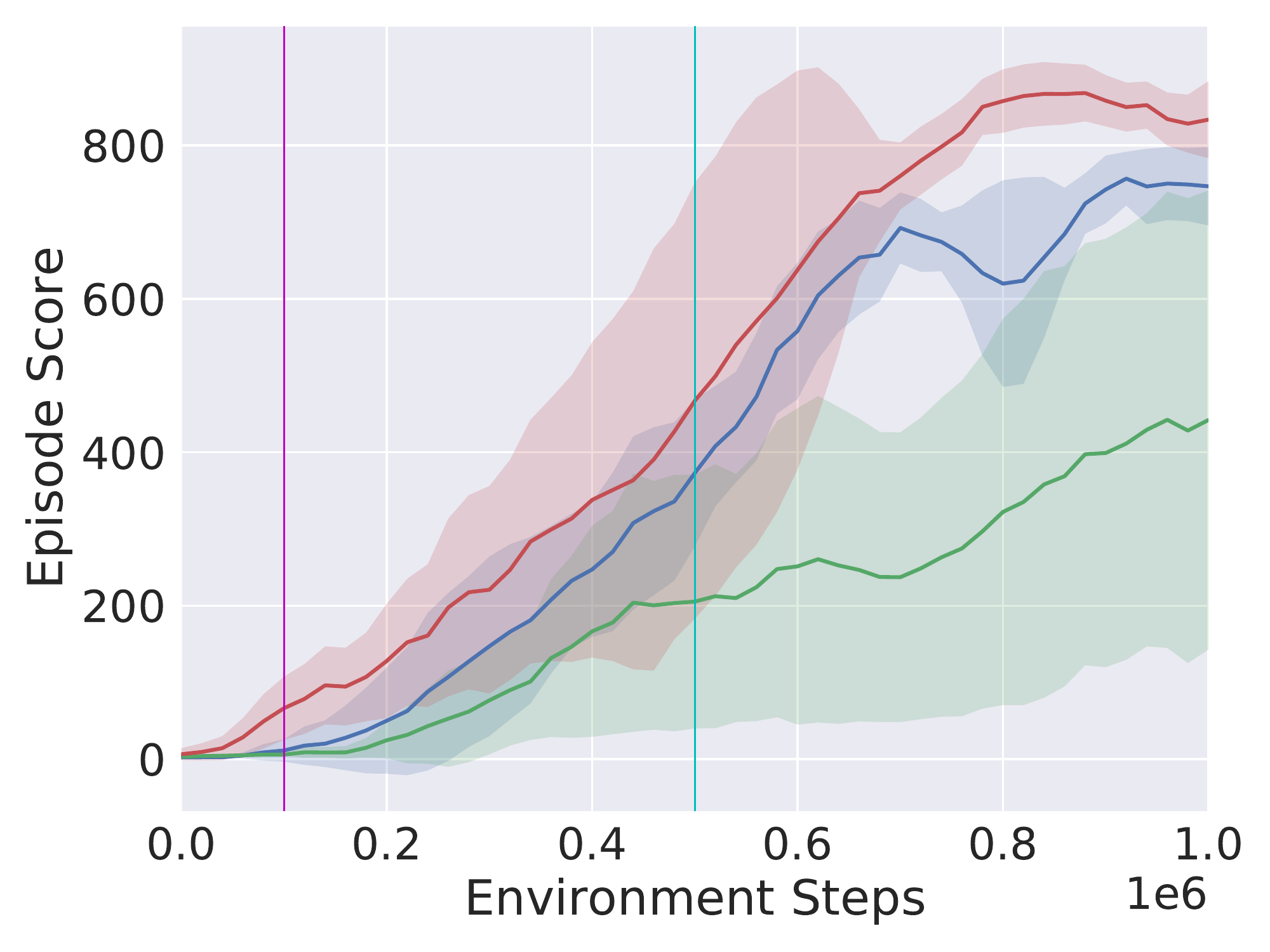}
    \caption{\small hopper-stand}
    \end{subfigure}
    \hfill
    \begin{subfigure}[b]{0.245\textwidth}
    \centering
    \includegraphics[width=\textwidth]{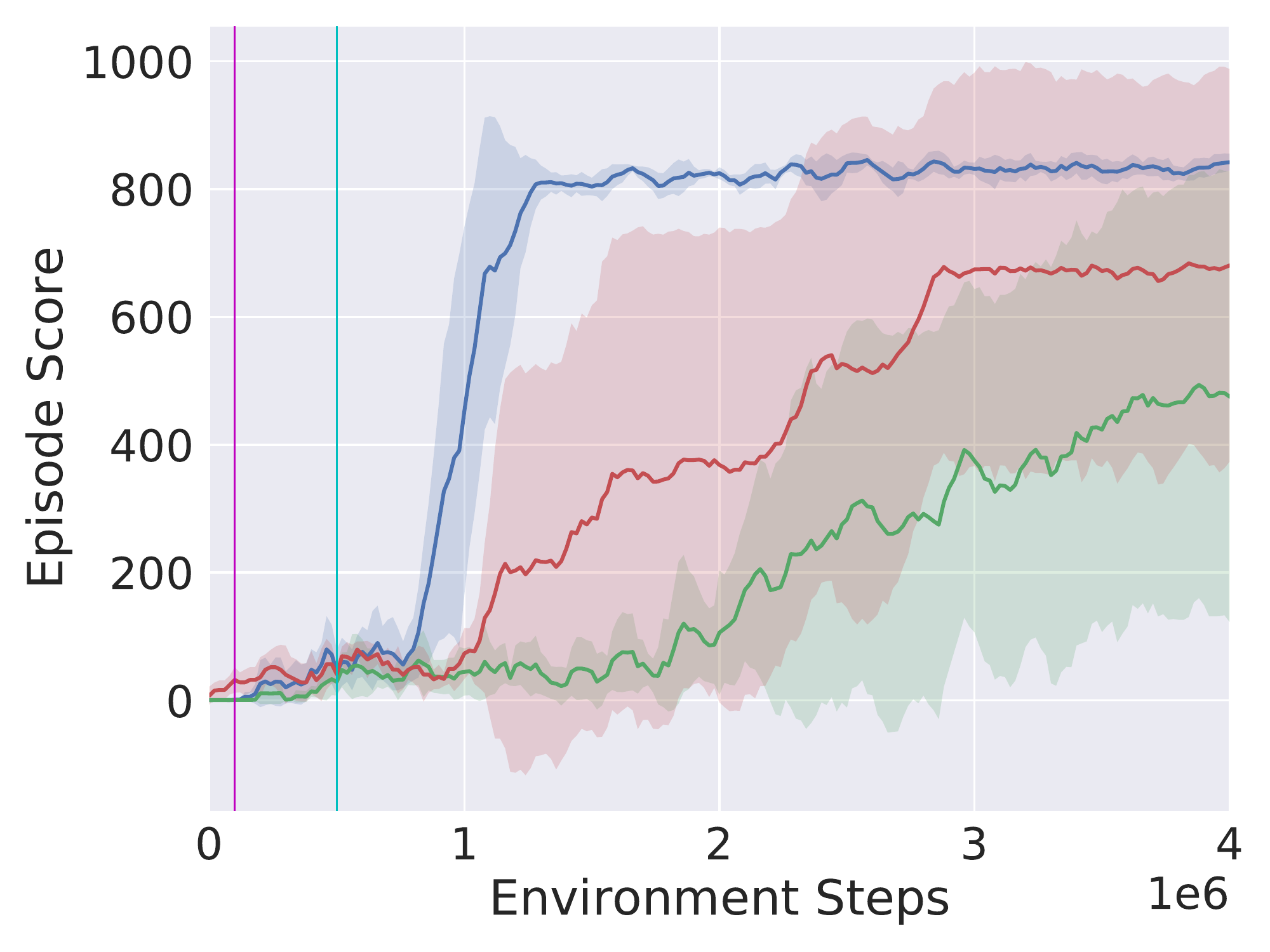}
    \caption{\small pendulum-swingup}
    \end{subfigure}

    \begin{subfigure}[b]{0.245\textwidth}
    \centering
    \includegraphics[width=\textwidth]{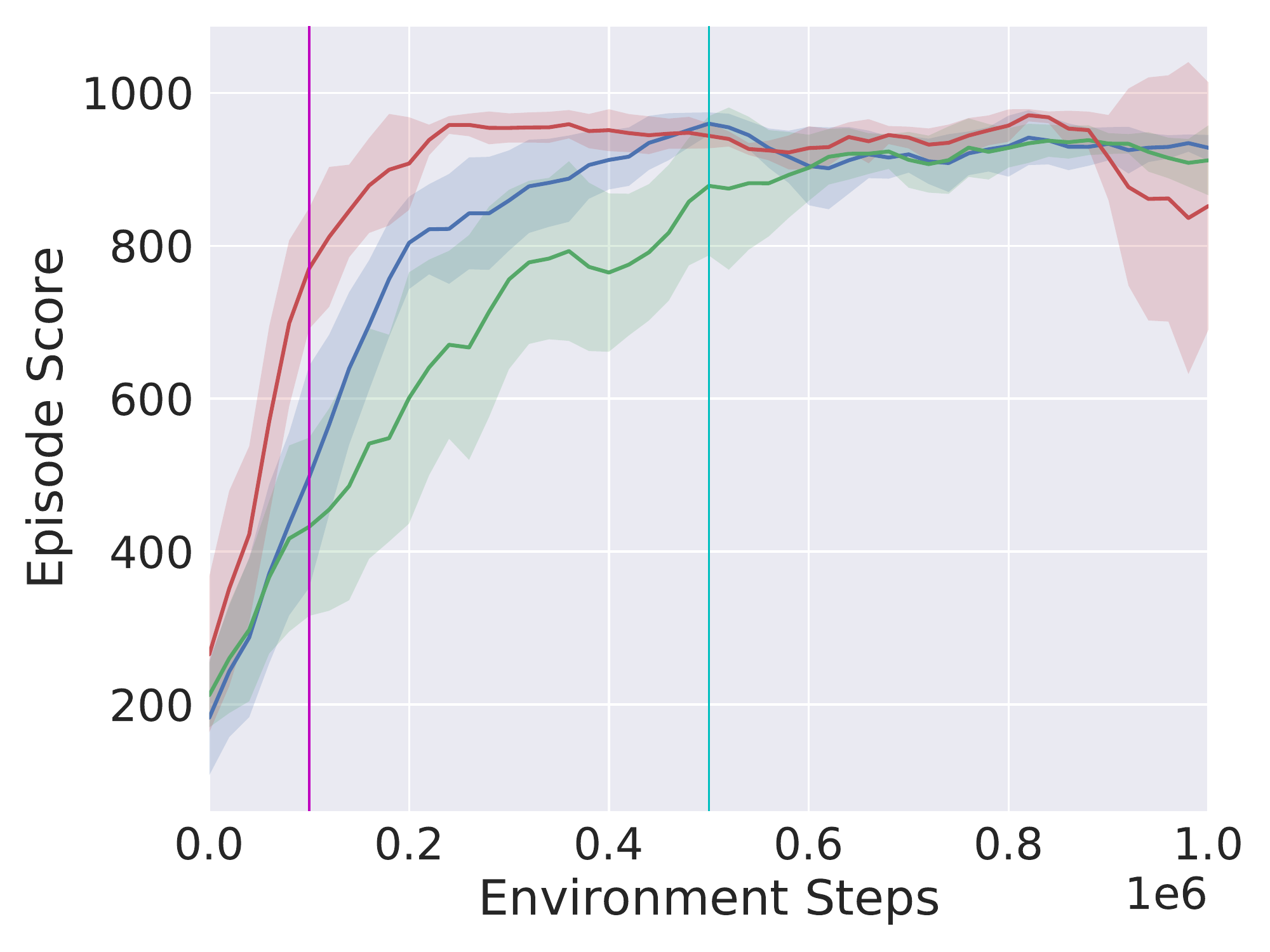}
    \caption{\small reacher-easy}
    \end{subfigure}
    \hfill
    \begin{subfigure}[b]{0.245\textwidth}
    \centering
    \includegraphics[width=\textwidth]{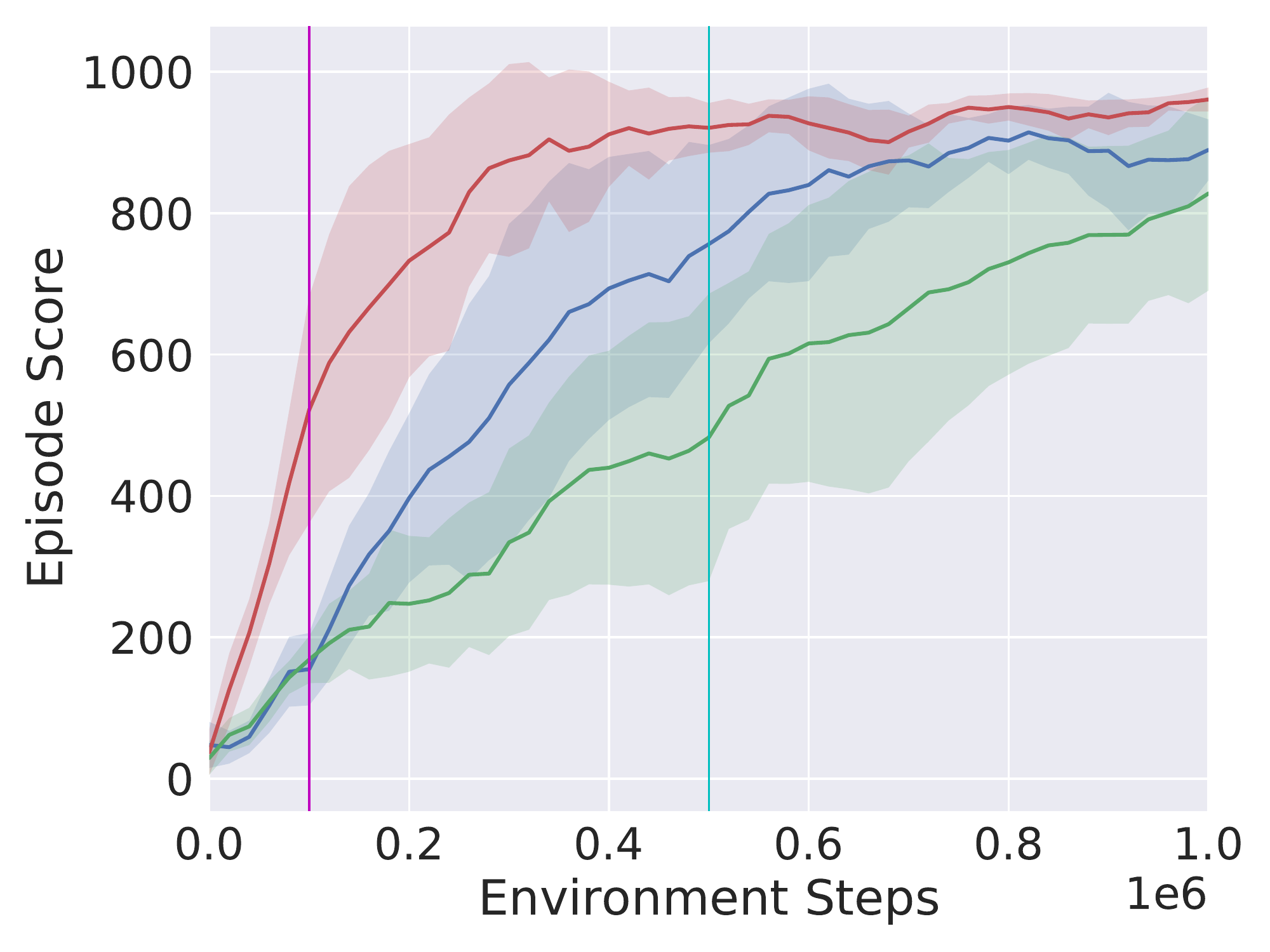}
    \caption{\small reacher-hard}
    \end{subfigure}
    \hfill
    \begin{subfigure}[b]{0.245\textwidth}
    \centering
    \includegraphics[width=\textwidth]{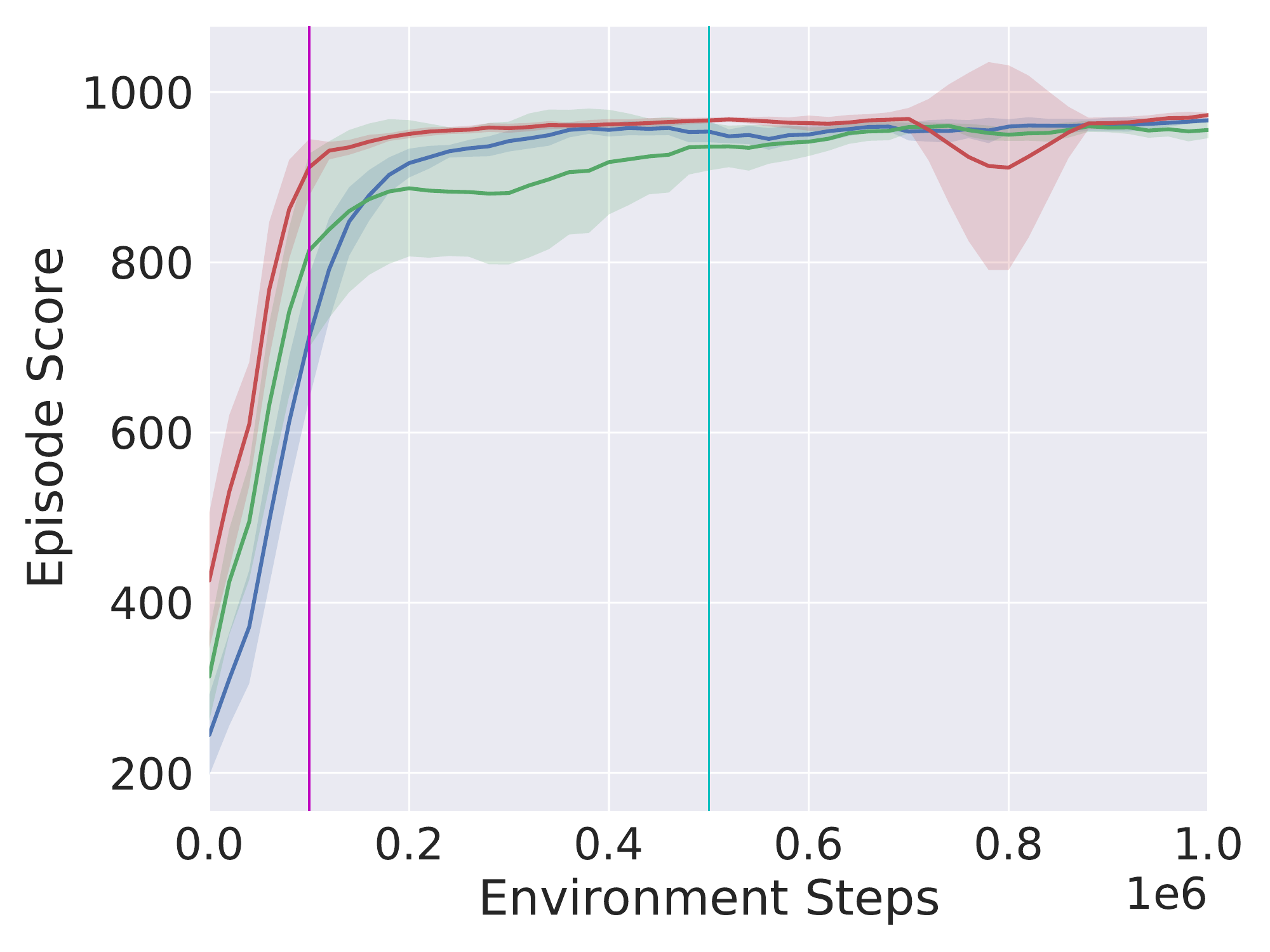}
    \caption{\small walker-stand}
    \end{subfigure}
    \hfill
    \begin{subfigure}[b]{0.245\textwidth}
    \centering
    \includegraphics[width=\textwidth]{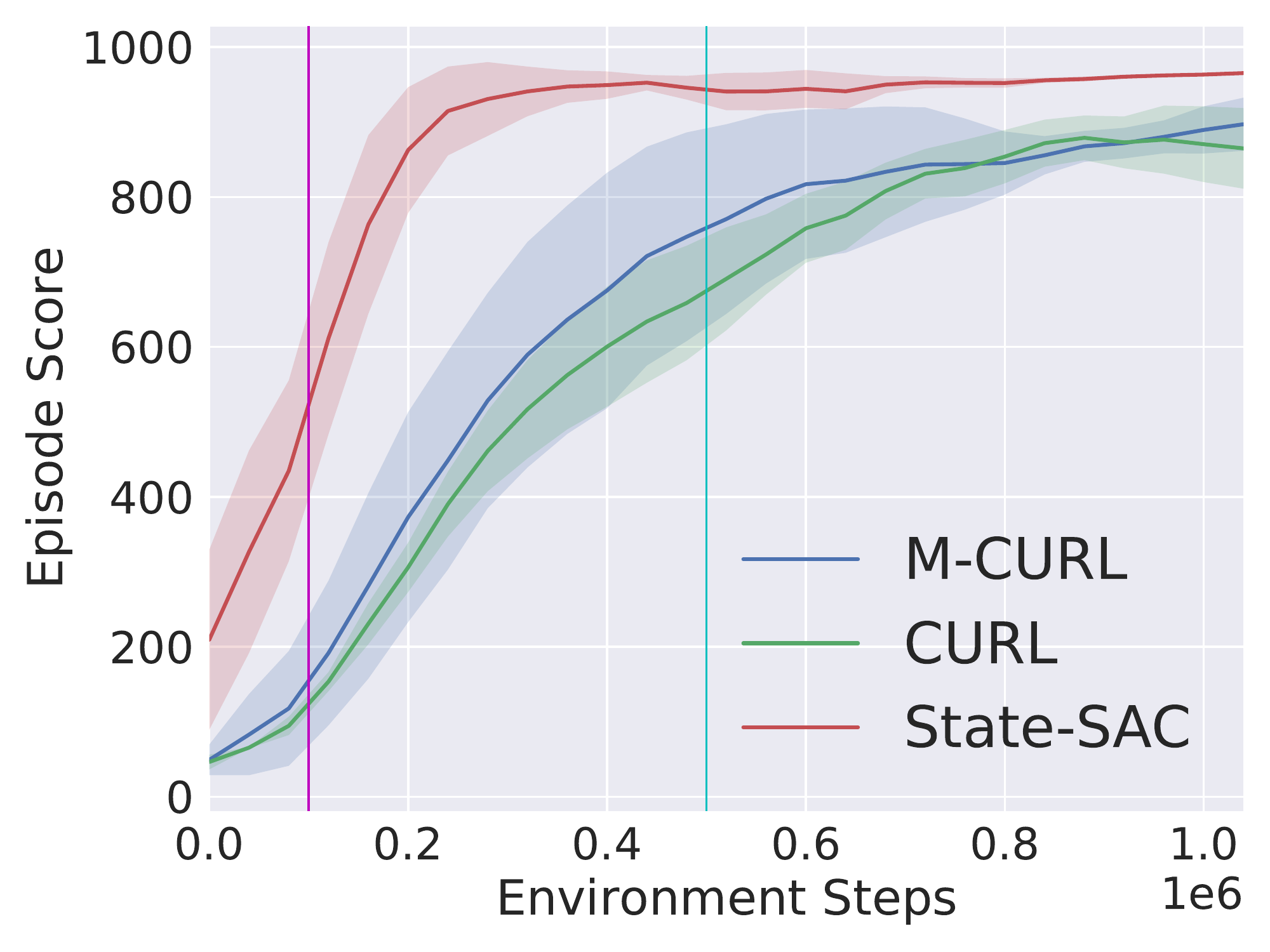}
    \caption{\small walker-walk}
    \end{subfigure}
    \vspace{-0.1cm}
    \caption{Results of CURL and our method on DMControl benchmark. Our method outperforms CURL on 14 out of 16 selected environments. The shaded region represents the standard deviation of the average return over 5 trials, and curves are smoothed uniformly for visual clarity. }
    \vspace{-0.4cm}
    \label{fig:dm_result}
\end{figure}

\section{Experiments}\label{sec:experiments}
In this section, we evaluate our method on two commonly used benchmarks, DMControl suite~\citep{tassa2020dmcontrol} and Atari 2600 Games~\citep{bellemare2013arcade}.

\subsection{Environments}
\noindent(1) {\it DMcontrol suite} contains a set of continuous control tasks 
powered by the MuJoCo physics engine~\citep{6386109}. 
These tasks are rendered through raw pixels which is suitable to test data-efficiency of different methods. Following the setup of CURL~\citep{srinivas2020curl}, we evaluate the performance at $100k$ (low sample regime) and $500k$ (asymptotically optimal regime) environment steps. We run experiments on sixteen environments from DMControl to evaluate the performance. Following common practice, we randomly crop the images rendered by game simulator from $100\times100$ to $84\times84$ for data augmentation. Action repeat, which refers to the number of times an action is repeated when it’s drawn from the agent’s policy, is set to 4 by default except that 2 for Finger Spin and Walker Walk, 8 for Cartpole Swingup. For each state, we stack the latest 3 observations as input, i.e., the $K$ in Section~\ref{sec:prelim} is $3$. 

\noindent(2) For discrete control tasks, we evaluate our method on {\it Atari 2600 games} from the arcade learning environment \citep{bellemare2013arcade}. We follow the training and evaluation procedures of \citet{srinivas2020curl}\footnote{\url{https://github.com/aravindsrinivas/curl_rainbow}}, to compare different algorithms in term of performance within 100k interaction steps (i.e., 400k environment step). We choose Rainbow~\citep{hessel2018rainbow} algorithm to train the policy network $\pi_\omega$. We run experiments on 26 atari games. Following common practice, the action repeat is set to 4 and frame stack is 4 across all games.

\begin{table}[!htb]
    \centering
   \scalebox{0.85}{
    \begin{tabular}{lccccccc}
    \toprule
GAME & HUMAN  & SIMPLE & OTRAINBOW & EFF. RAINBOW & CURL & \ourM \\
\midrule
ALIEN & $7127.7$  & $616.9$ & $824.7$ & $739.9$ & $558.2$&$\mathbf{ 847.4}$ \\
AMIDAR & 1719.5   & 88.0 & 82.8 & $\mathbf{188.6}$ & 142.1 & 168.0 \\
ASSAULT & 742.0   & 527.2 & 351.9 & 431.2 & 600.6  & $\mathbf{606.8}$\\
ASTERIX & 8503.3  & $\mathbf{1128.3}$ & 628.5 & 470.8 & 734.5 &629.0 \\
BANK HEIST & 753.1   & 34.2 & $\mathbf{182.1}$ & 51.0 & 131.6 & 176.6\\
BATTLE ZONE & 37187.5   & 5184.4 & 4060.6 & 10124.6 & 14870.0& $\mathbf{21820.0}$ \\
BOXING & 12.1   & $\mathbf{9 . 1}$ & 2.5 & 0.2 & 1.2 &3.5 \\
BREAKOUT & 30.5   & $\mathbf{1 6 . 4}$ & 9.84 & 1.9 & 4.9 &3.5\\
CHOPPER COMMAND & 7387.8  & $\mathbf{1 2 4 6 . 9}$ & 1033.33 & 861.8 & 1058.5 & 1096.0 \\
CRAZY\_CLIMBER & 35829.4  & $\mathbf{6 2 5 8 3 . 6}$ & 21327.8 & 16185.3 & 12146.5&19420.0 \\
DEMON\_ATTACK & 1971.0   & 208.1 & 711.8 & 508.0 & 817.6&$\mathbf{860.8}$ \\
FREEWAY & 29.6  & 20.3 & 25.0 & $27.9$ & 26.7&$\mathbf{28.6}$ \\
FROSTBITE & 4334.7   & 254.7 & 231.6 & 866.8 & 1181.3&$\mathbf{2154.8}$ \\
GOPHER & 2412.5  & 771.0 & $\mathbf{7 7 8 . 0}$ & 349.5 & 669.3 & 426.4 \\
HERO & 30826.4   & 2656.6 & 6458.8 & 6857.0 & 6279.3& $\mathbf{6884.9}$ \\
JAMESBOND & 302.8  & 125.3 & 112.3 & 301.6 & $\mathbf{4 7 1 . 0}$ &368.0\\
KANGAROO & 3035.0   & 323.1 & 605.4 & 779.3 & 872.5 &$\mathbf{1410.0}$ \\
KRULL & 2665.5   & $\mathbf{4 5 3 9 . 9}$ & 3277.9 & 2851.5 & 4229.6 &2928.9\\
KUNG\_FU\_MASTER & 22736.3   & $\mathbf{1 7 2 5 7 . 2}$ & 5722.2 & 14346.1 & 14307.8 & 15804.0 \\
MS\_PACMAN & 6951.6   & 1480.0 & 941.9 & 1204.1 & 1465.5&$\mathbf{1687.2}$ \\
PONG & 14.6 & $\mathbf{12.8}$ & 1.3 & -19.3 & -16.5 &-16.8\\
PRIVATE EYE & 69571.3  & 58.3 & 100.0 & 97.8 &218.4 &$\mathbf{293.4}$ \\
QBERT & 13455.0 & $\mathbf{1 2 8 8 . 8}$ & 509.3 & 1152.9 & 1042.4 & 1272.5 \\
ROAD\_RUNNER & 7845.0 & 5640.6 & 2696.7 & $\mathbf{9 6 0 0 . 0}$ & 5661.0 & 8890.0 \\
SEAQUEST & 42054.7   & $\mathbf{6 8 3 . 3}$ & 286.92 & 354.1 & 384.5 & 444.4 \\
UP\_N\_DOWN & 11693.2  & 3350.3 & 2847.6 & 2877.4 & 2955.2&$\mathbf{3475.6}$ \\
\midrule
Median HNS & $100.0\%$ & $14.4\%$ &$20.4\%$ & $16.1\%$ & $17.5\%$ &$\mathbf{24.2\%}$ \\
    \bottomrule
    \end{tabular}}
    \caption{Results of our method and baselines on 26 environments from Atari 2600 Games at 100k timesteps. Our method achieves state-of-the-art performance on 10 out of 26 environments. Our method is implemented on top of Eff. Rainbow \citep{van2019use}, improves its performance on 24 out of 26 environments, and outperforms CURL on 21 out of 26 environments. The scores are averaged over 5 different seeds, and we also can see our method achieves superhuman performance on JamesBond, Krull and Road\_Runner. }
    \vspace{-0.3cm}
    \label{tab:atari}
\end{table}

\subsection{Baselines}
For DMControl Suite, CURL~\citep{srinivas2020curl} has demonstrated its superiority over previous methods designed for sample efficiency, such as PlaNet~\citep{hafner2019learning}, Dreamer~\citep{hafner2019dream}, SAC+AE \citep{yarats2019improving}, SLACv1~\citep{lee2019stochastic}. Due to computation limitation, we compare our method with CURL and state-SAC. With state-SAC, the agents can access to the low level states of the 16 selected DMControl environments, such as positions
and velocities. State-SAC is usually implemented as a baseline for the algorithms about improving data efficiency, which can be used to measure the gap between the proposed method and the approximate upper-bound scores which pixel-based agents can achieve. Each algorithm is run with five different seeds.  



For Atari 2600 Games, we compare our method with the following baselines: (i) Human performance~\citep{kaiser2019model,van2019use}; 
(ii) SimPLe \citep{kaiser2019model}, the best model-based algorithm for Atari; (iii) OTRainbow \citep{kielak2020recent}, the over-trained version of Rainbow, (iv) Efficient Rainbow \citep{van2019use}, which modifies the the hyper-parameters and architecture of Rainbow for data efficiency, (v) CURL \citep{srinivas2020curl}, which first integrates contrastive learning in Rainbow for data efficiency. Following \citet{srinivas2020curl}, our method is implemented on top of Efficient Rainbow.

\subsection{Model Configuration}
The architecture of the CNN encoder $f_\theta$ is the same as that of CURL \citep{srinivas2020curl} for a fair comparison. The momentum coefficient is $0.05$ for DMcontrol and $0.001$ for Atari game. We use one 2-layer Transformer encoder with single-head attention for DMControl and Atari benchmark. The masked percentage in data preparation $\varrho_m$ is $0.6$ for Finger Spin and Walker Walk, $0.5$ for others. The other hyperparameters remain the same as~\citep{srinivas2020curl}\footnote{\url{https://github.com/MishaLaskin/curl}}. The two objects $\mathcal{L}_{rl}$ and $\mathcal{L}_{cl}$ are weighted equally, and updated jointly per interaction step. We use Adam \citep{kingma2014adam} optimizer for training. The detailed parameters are left in Appendix \ref{hyperpara}.

\subsection{Results}
The results of DMControl are shown in Figure \ref{fig:dm_result}. Generally, our method outperforms the standard CURL on 14 out of 16 experiments, and is slightly worse than CURL on the remaining two (Cartpole-Balance-Sparse and Walker-Stand), which demonstrates the effectiveness of our method. We have the following observations:


(1) From the perspective of sample efficiency, on most of these tasks, the slopes of the curves of our method are much larger than those of CURL. That is, with the same amount of interaction with environments, our method can achieve better performance. As shown in the figures, when the interaction steps are $100k$ (magenta vertical line) and $500k$ (cyan vertical line), our method achieves significantly better results than CURL. On some difficult tasks like Cartpole-Swingup-Sparse and Pendulum Swingup, our proposed \ourM{}~requires much fewer steps than CURL to get positive rewards. These results demonstrate that our method is more sample efficient.


(2) Our method further reduces the gap between CURL and state-SAC, which is trained from internal state rather than from pixel. State-SAC is expected to be an approximate upper bound of algorithms for improving data efficiency. We can even observe that in Cheetah-Run, Finger-Spin, Hopper-Hop and Pendulum-Swingup tasks, our method achieves higher scores than state-SAC, which again shows the effectiveness of our method.

We also conduct experiments on Atari with $100k$ interaction steps. The results are shown in Table \ref{tab:atari}. Again, our method achieves better score than CURL. Specifically,

(1) Our method \ourM{} surpasses CURL on $21$ out of $26$ games, which shows that \ourM{} can bring improvements to discrete control problems. Our method outperforms Efficient Rainbow DQN on $24$ out of $26$ Atari games, which is the algorithm that \ourM{} built on top of. This shows the generality of our method. Besides, \ourM{} achieves super-human performances on JamesBond, Krull and Road\_Runner by $24\%$, $25\%$ and $13\%$ improvements, which shows the effectiveness of our method.

(2) We also provide the median human-normalized score (HNS) of all algorithms, which is calculated as follows: Given $M$ games, we first get the human scores and the scores of a random algorithm, denoted as $S_{H,i}$ and $S_{R,i}$ respectively, $i\in[M]$. For any algorithm $A$, the score of the $i$-th game is $S_{A,i}$. Medium HNS is the medium value of $\{\frac{S_{A,i}-S_{R,i}}{S_{H,i}-S_{R,i}}|i\in[M]\}$. The median HNS of \ourM{} is $24.2\%$, while the meadian HNS of CURL, SimPLe and Efficient Rainbow DQN are $17.5\%$, $14.4\%$ and $16.1\%$ respectively, which shows that in general, our method achieves the best performance across different tasks. 


\section{Ablation Study}\label{sec:ablation}
To better understand our framework, we provide several further study experiments and analyses.

\subsection{Study on masked probability}
\begin{figure}[!htpb]
    \centering
    \begin{minipage}{0.46\textwidth}
    \begin{subfigure}[h]{\textwidth}
    \centering
    \includegraphics[width=\textwidth]{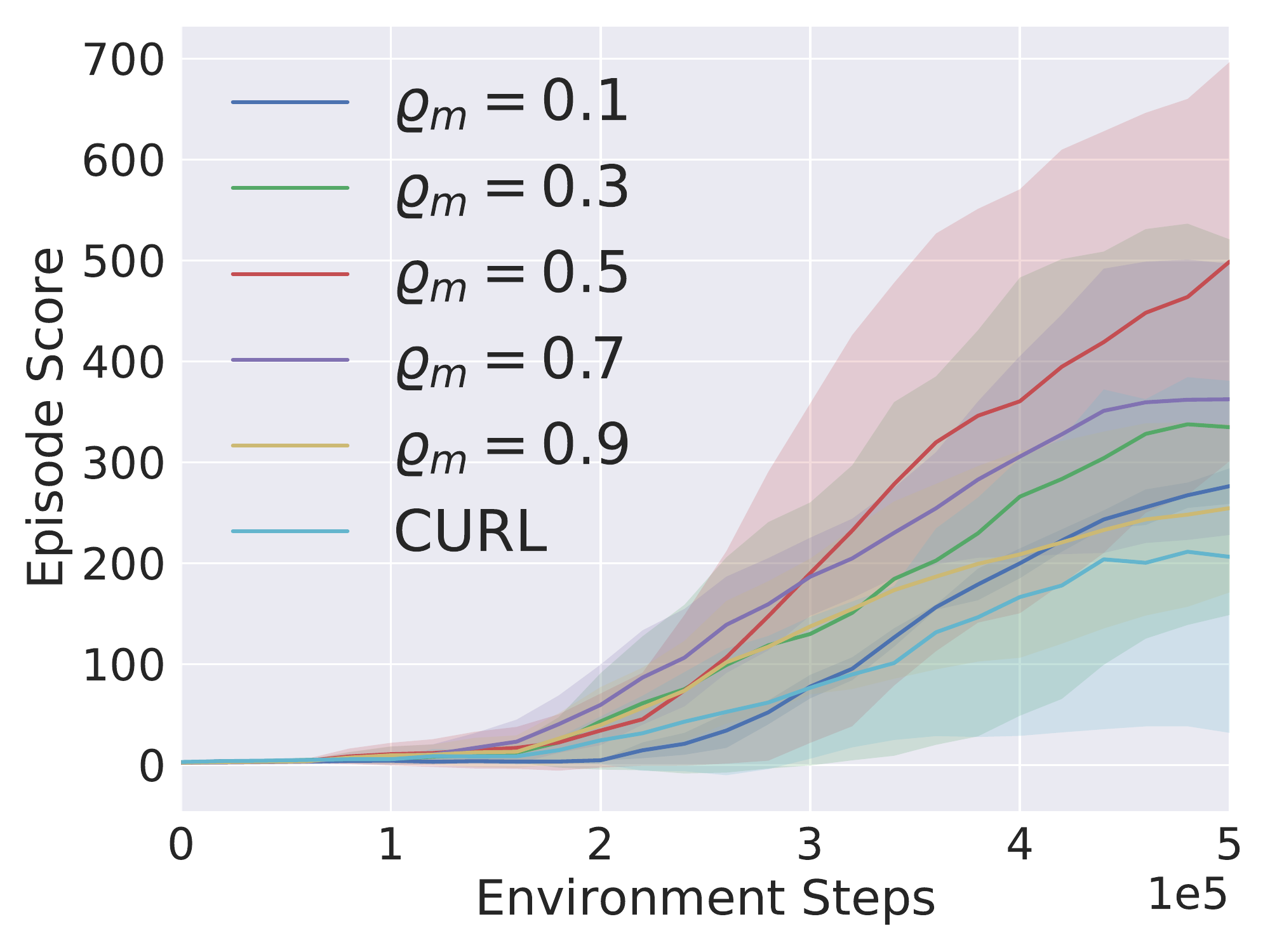}
    \caption{Scores on Hopper-Stand environment achieved by different masked probability $\varrho_m$.}
    \label{fig:ratio}
    \end{subfigure}
    \end{minipage}
    \begin{minipage}{0.46\textwidth}
    \vspace{-0.35cm}
    \begin{subfigure}[h]{\textwidth}
    \centering
    \includegraphics[width=\textwidth]{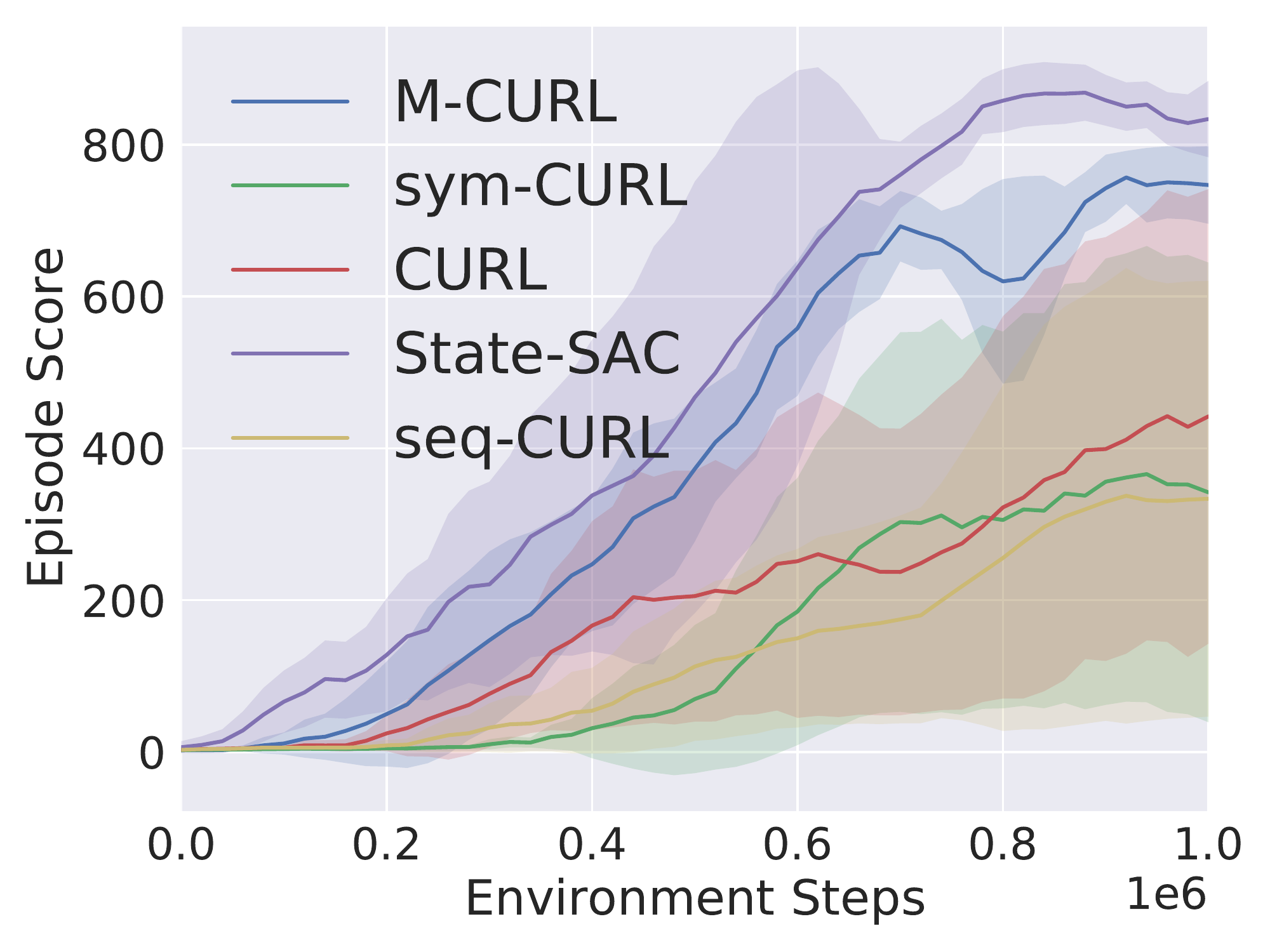}
    \caption{Comparison with different variants of CURL.}
    \label{fig:seq}
    \end{subfigure}
    \end{minipage}
    \vspace{-0.2cm}
    \caption{Results of ablation study.}
    \vspace{-0.2cm}
    \label{fig:ablation}
\end{figure}

We explore the effect of different masked probability $\varrho_m$ in this section.
We conduct experiments on Hopper-Stand environment with $\varrho_m \in \{ 0.1, 0.3, 0.5, 0.7, 0.9 \}$. The results are shown in Figure~\ref{fig:ratio}. Our method achieves better results than standard CURL w.r.t. different $\varrho_m$, which demonstrates the effectiveness of using masked training.


As shown in Figure~\ref{fig:ratio}, sample-efficiency indeed varies w.r.t. $\varrho_m$. (1) With $500k$ interactions, setting $\varrho_m$ as $0.5$ achieves the best performance. (2) Setting a too large $\varrho_m$ (i.e., $0.9$) will not lead to good results, since too much information is masked and we cannot reconstruct a reasonable feature. Also, setting a too small $\varrho_m$ (i.e., $0.1$) will hurt the performance too, because there are fewer positions to provide the training signal (see \eqref{eq:contrastive_loss_}), and leveraging more information will reduce the ability of remembering past information and predicting future information.


\subsection{Study on Variants of CURL}

\noindent(1) {\bf CURL with consecutive inputs}

With our method, the input of the Transformer module is consecutive $T$ observations from $\mathcal{S}$, which is then processed into query, key and negative samples for contrastive learning. However, in standard CURL, these $T$ frames are randomly generated to simulate an i.i.d. data generation process~\citep{mnih2013playing,haarnoja2018soft}. To investigate whether the improvements of our method are from the usage of consecutive inputs only, we implement a variant of CURL (denoted as seq-CURL), where the inputs are not randomly generated from replay buffer, but sequentially selected from the buffer (see Appendix \ref{pseudo-code-seqcurl} for pseudo-code). The results are in Figure~\ref{fig:seq}. Compared to standard CURL, the results become even worse. This shows the importance of using randomly sampling in CURL, and demonstrates that the improvement of our method is not simply due to the usage of consecutive inputs. 

\noindent(2) {\bf M-CURL with two Transformer modules}

In our method, the queries are encoded by a Transformer module while the keys are not. A natural baseline is that the keys should also be encoded using a Transformer module. We implement this variant (denoted as sym-CURL), where the Transformer module of keys are updated via momentum contrastive learning. The results are in Figure~\ref{fig:seq}. We can see that using a symmetric architecture hurts the performance too and the absence of Transformer module above the momentum encoder is crucial in our method. Our conjecture is that in \ourM{}, the contrastive loss is directly applied to the features output by the CNN encoder, which is the inputs of policy network, and thus force it capture the correlation between consecutive observations like the outputs of Transformer.




Additional ablation studies are left in Appendix \ref{sec:more_ablation_study} due to space limitations, including the study on the consecutive length $T$ and the study of Transformer module with different numbers of layers. 

\section{Conclusion and Future Work}\label{sec:conclusion}
In this work, we proposed \ourM{}, which leverages masked training and Transformer module to improve sample efficiency for reinforcement learning. Experiments on DMControl and Atari benchmark demonstrate the effectiveness of our method and show the benefit of utilizing the correlation among the adjacent video frames. For future work, there are many directions to explore. 

First, how to efficiently adapt our idea into on-policy RL algorithm is an interesting topic. Second, how to leverage both the states and actions (rather than only states in this work) saved in replay buffer into our method remains to be explored. Third, we will study how to use more advanced models and pre-training techniques to improve sample-efficiency for RL.

\bibliography{iclr2021_conference}
\bibliographystyle{iclr2021_conference}

\appendix
\section{More Details about Experiment Setup}\label{hyperpara}
\subsection{Network architecture}

\subsubsection{Encoder network}
For DMControl, we use the encoder architecture from \citet{yarats2019improving}, which is the same as that in \citet{srinivas2020curl}. The encoder is stacked by four convolutional layers with $3\times3$ kernel size and 32 channels, and one fully-connected layer which outputs 50-dimension features. Each convolutional layer is followed by a \texttt{ReLU} activation and the output of fully-connected layer is normalized by \texttt{LayerNorm} \citep{ba2016layer}. For Atari, we use the data-efficient version of Rainbow DQN \citep{van2019use}, and the encoder is stacked by two convolutional layers with $5\times5$ kernel size and 32, 64 channels respectively. These two convolutional layers are connected by \texttt{ReLU} activation.

\subsubsection{Actor and Critic networks}
For DMcontrol, we use the SAC \citep{haarnoja2018soft} algorithm, and there are a critic network and an actor network. The critic network consists of 3 fully connected layers. Each fully-connected layer (except the last one) is followed by a \texttt{ReLU} activation. The actor network also consists of 3 fully connected layers, which output the mean values and covariance for one diagonal Gaussian distribution to reparameterize the policy. For Atari, we use the efficient Rainbow DQN \citep{van2019use}, and there is only one Q-network consisting of a value branch and an advantage branch. Each branch is 2-layers MLP which intermediated by the \texttt{ReLU} activation, and we use Noisy Network \citep{fortunato2017noisy} for exploration.

\subsubsection{Positional Embedding} The outputs of the image encoder $f_\theta$ do not contain positional information, which cannot model the temporal information among different inputs. To overcome this difficulty, following~\citet{vaswani2017attention}, we introduce positional embedding into the Transformer module, which is a $d$-dimensional vector and not to be learned. 
For each position $k$, denote the positional embedding as $p_k\in\mathbb{R}^d$. Note that in Section 3.2 (line 233), we also denote the image representation as a $d$-dimensional vector. Let $p_{(k,j)}$ denote the $j$-th element of $p_k$, $j\in[d]$. We have that
\begin{equation}\begin{aligned}
p_{(k, 2 i)} &=\sin \left(k / 10000^{2 i / d}\right), \\
p_{(k, 2 i+1)} &=\cos \left(k / 10000^{2 i / d}\right),\\
\end{aligned}\end{equation}
where $2i$ and $2i+1$ are the dimension indicators. The augmented query representation of position $k$ becomes
\begin{equation}
h_k = f_\theta(\bar{s}^\prime_k) + p_k.
\end{equation}

\subsubsection{Transformer Encoder Block}\label{transformer_arch}
We then process these query representations (obtained by using positional embedding) with several stacked Transformer encoder blocks. Each block consists of two layers, namely a self-attention layer $\operatorname{attn(\cdots)}$ and a feed-forward layer $\operatorname{ffn(\cdots)}$. Specifically, let $h^l_i$ denote the output of $i$-th position of the $l$-th block, and $H^l=(h^l_1,h^l_2,\cdots,h^l_T)$. For any $l\in[L]$, we have that

\begin{equation}\begin{aligned}
\alpha_{ij}&=\frac{\exp \left(\left(W_{q} h^{l-1}_i\right)^{T}\left(W_{k} h^{l-1}_{j}\right)\right)}{Z},\\ 
Z&=\sum_{j=1}^{|K|} \exp \left(\left(W_{q} h^{l-1}_i\right)^{T}\left(W_{k} h^{l-1}_{j}\right)\right), \\
\hat{h^l_i} &=\operatorname{attn}(h^{l-1}_i, H^{l-1}, H^{l-1})=\sum_{j=1}^{|V|} \alpha_{ij} W_{v} h^{l-1}_{j},\\ 
\tilde{h}^l_{i} &= \operatorname{LayerNorm}(h^{l-1}_i + \hat{h^l_i}),\\
\bar{h}^l_i &= \operatorname{ffn}(\tilde{h}^l)= W_{2} \max \left(W_{1} \tilde{h}^l_i+b_{1}, 0\right)+b_{2},\\
h^l_{i} &= \operatorname{LayerNorm}(\tilde{h}^l_{i} + \bar{h}^l_{i}),
\end{aligned}\end{equation}
where the $W_q, W_k, W_v, W_1, W_2$ are learnable parameters.    
After $L$ encoder block, we obtain $H^L$ as the query set $Q$ for contrastive learning.

\subsection{Learning Rate Scheduler}\label{lrs}
Specifically, following \citet{vaswani2017attention}, the learning rate scheduler of the Transformer module is defined as follows:
\begin{equation}
\eta = \eta_0 \times \min \left((\frac{\operatorname{step}}{\operatorname{step}_w})^{-0.5},\frac{\operatorname {step}}{\operatorname{step}_w}\right),
\end{equation}
where $\eta_0$ is the base learning rate which is the same as that used in optimizing RL obeject, $\operatorname{step}$ is the current training step and $\operatorname{step}_w$ is the warmup step.  We set $\operatorname{step}_w$ as $6000$ in our experiments. 

\subsection{Other Hyper-parameters}
We mainly follow the setup used in \citet{srinivas2020curl}, and the other hyper-parameters are summarized below. 
For DMControl benchmark,  the replay buffer size is $100k$, and the batchsize is 512 for cheetah and 128 for others. We use Adam \citep{kingma2014adam} to optimize all our parameters with $\beta_1=0.9,\beta_2=0.999$ for all our network parameters and $\beta_1=0.5, \beta_2=0.999$ for the temperature coefficient in SAC. The learning rate is $1e-4$ for all optimizer and we use \texttt{inverse\_sqrt} learning rate scheduler described in \ref{lrs} for Transformer module. For Atari benchmark, the batchsize is 32 and the multi-step return length is 20. The optimizer settings and learning rates are the same as those used in DMControl.
\section{More Ablation Study}\label{sec:more_ablation_study}
\subsection{Study on Sequence Length}
\begin{figure}[!htpb]
    \centering
    \begin{minipage}{0.46\textwidth}
    \begin{subfigure}[h]{\textwidth}
    \centering
\includegraphics[width=\textwidth]{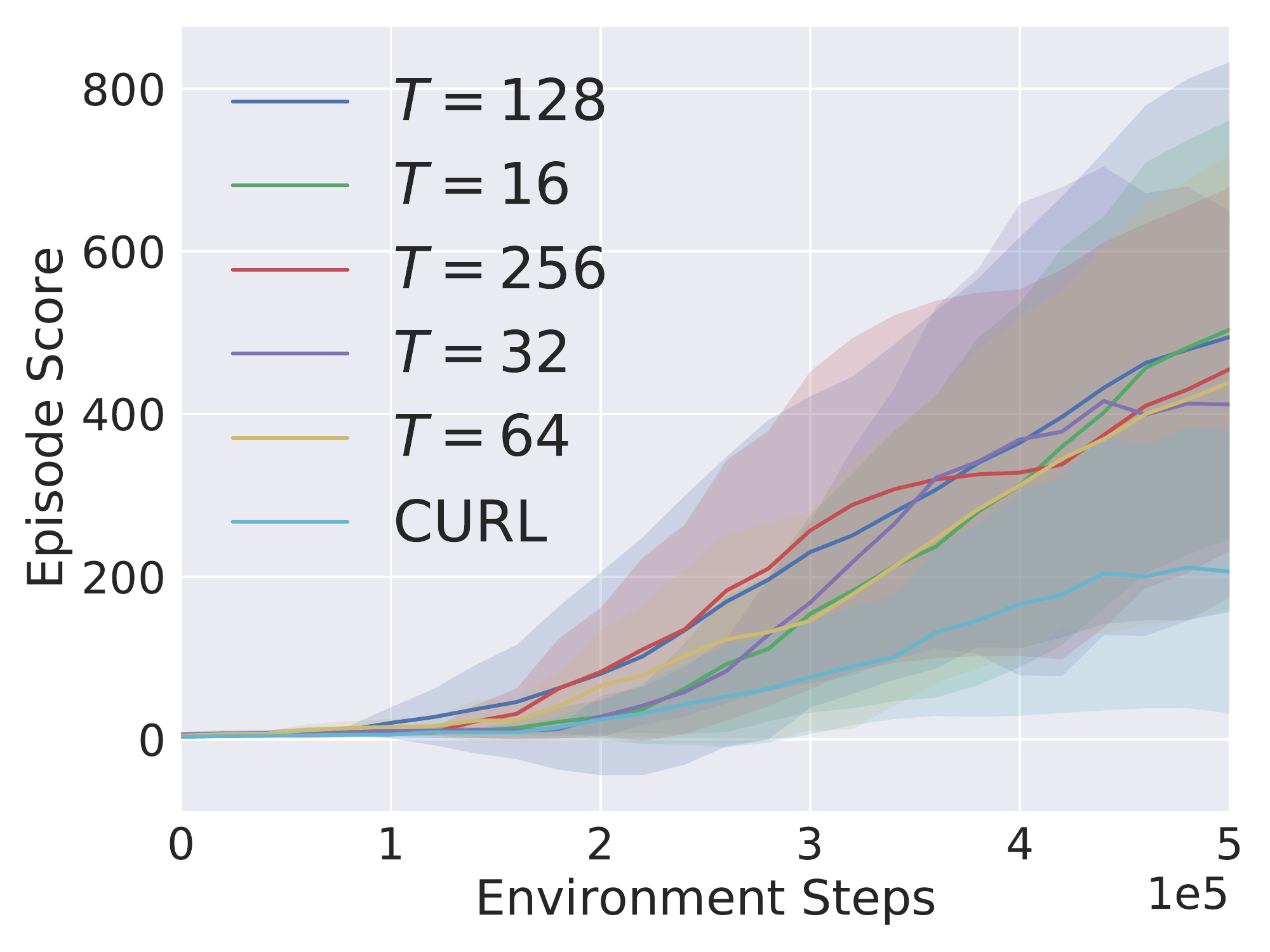}
    \caption{Scores on Hopper-Stand environment achieved by different sequence length.}
    \label{fig:length}
    \end{subfigure}
    \end{minipage}
    \begin{minipage}{0.46\textwidth}
    \vspace{-0.35cm}
    \begin{subfigure}[h]{\textwidth}
    \centering
        \includegraphics[width=\textwidth]{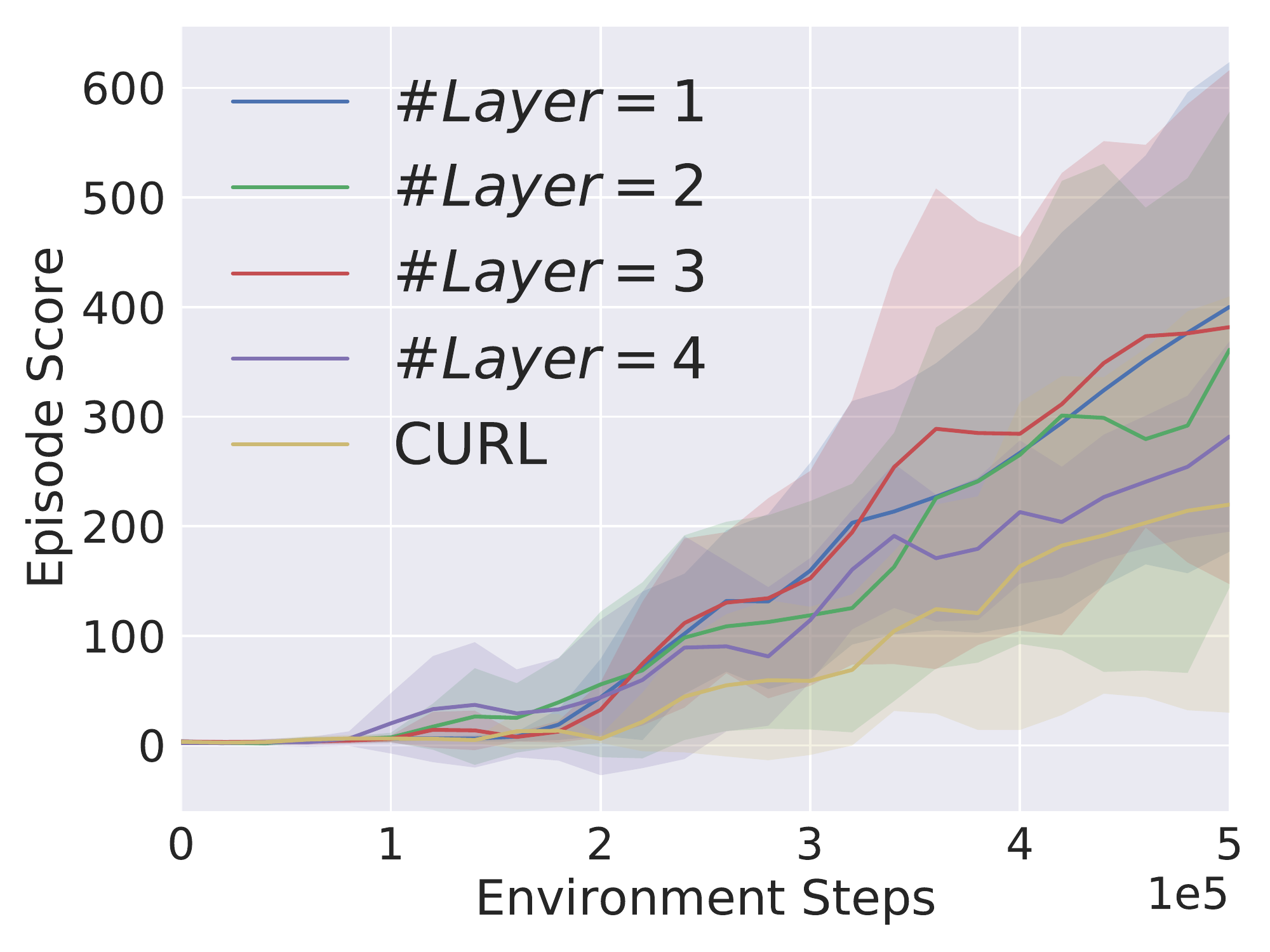}
    \caption{Scores on Hopper-Stand environment achieved by different layer number.}
    \label{fig:layernumer}
    \end{subfigure}
    \end{minipage}
    \vspace{-0.2cm}
    \caption{Results of ablation study.}
    \vspace{-0.2cm}
    \label{fig:moreablation}
\end{figure}

The role of Transformer module we used in our method is to capture temporal correlation in state sequences. The sequence length $T$ is also another key factor for learning the  representations.
In this section, to investigate the effect of different sequence lengths, we conduct experiments on Hopper-Stand environment with different sequence length, $T \in \{ 16, 32, 64, 128, 256 \}$, where we change the corresponding batch size $B$ to ensure the same number of training samples $T\times B$ in each parameter update. The results are shown in Figure \ref{fig:length}.

As shown in the figure, we find that there is no much difference between the scores achieved by different sequence lengths, which shows the our method is not sensitive to the choice of $T$. The result of length 16 is competitive with other results of longer lengths, which means that the effective context to reconstruct one state representation in RL settings is much smaller than that in NLP problems which is always hundreds of tokens. 

\subsection{Study on the Layer Number}
We also vary the number of layers of the Transformer module in $\{1, 2, 3, 4\}$. The result is shown in Figure \ref{fig:layernumer}. We notice that when the layer numbers are set as $\{1,2,3\}$, there is not significant difference. When the number of layer is increased to four, the performance slightly drops. We can also observe that the all the four settings outperform the standard CURL baseline, which demonstrates the effectiveness of our method.


\section{Algorithm}\label{algo_sac}

\begin{algorithm}[H]
  \footnotesize
  \caption{Masked Contrastive Representation Learning for Reinforcement Learning (\ourM)  coupling with SAC}
  \label{alg:mcurl}
  \begin{algorithmic}[1]
    \Require Environment $E$, initial parameters $\theta$, $\phi$, $\omega_1$, $\omega_{21}$, $\omega_{22}$ for the convoluational encoder, Transformer module, actor network $\pi_{\omega_1}$, and two critics $Q_{\omega_{21}}, Q_{\omega_{22}}$, and an empty replay buffer $\mathcal{B}$.
    \State $\bar{\theta} \leftarrow \theta$
    \State $\bar{\omega}_{2i} \leftarrow \omega_{2i}$ for $i \in \{1, 2\}$
    \State $s_{1} \sim E_{\text{reset}}()$  %
    \For{each iteration $t$ }
      \For{each environment step}
           \State $a_t \sim \pi_{\omega_1}(\cdot|s_t)$
           \State $s_{t+1}, r_t, d_t \sim E$
           \State $\mathcal{B} \leftarrow \mathcal{B} \cup (s_t, a_t, s_{t+1}, r_t, d_t)$
      \EndFor
      \For{each gradient step}
          \State $x_{1:T+1}, a_{1:T}, r_{1:T} \sim \mathcal{B}$ \Comment{i.i.d samples}
          \State $\omega_{2i} \leftarrow \omega_{2i} - \lambda_Q \nabla_{\omega_{2i}}J_Q(x_{1:T+1}, a_{1:T}, r_{1:T};\omega_{2i})$, for $i \in \{1,2\}$
          \State $\theta \leftarrow \theta - \lambda_\theta \nabla_\theta J_Q(x_{1:T+1}, a_{1:T}, r_{1:T};\theta)$, for $i \in \{1,2\}$
          \State $\omega_1 \leftarrow \omega_{1} - \lambda_\pi \nabla_{\omega_1}J_{\pi}(x_{1:T+1}, a_{1:T}, r_{1:T};\omega_1)$
          \State $x_{1:T} \sim \mathcal{B}$ \Comment{consecutive samples}
          \State $\phi \leftarrow \phi - \lambda_\phi \nabla_\phi J_{cl}(x_{1:T;\phi}) $
          \State $\theta \leftarrow \theta - \lambda_\theta \nabla_\theta J_{cl}(x_{1:T;\theta})$
          \State $\bar{\theta} \leftarrow \tau_1 \theta + (1-\tau_1)\bar{\theta} $ 
          \State $\bar{\omega}_{2i}\leftarrow \tau_2 \omega_{2i} + (1-\tau_2) \bar{\omega}_{2i} $, for $i \in \{1, 2\}$ 
      \EndFor
    \EndFor
  \end{algorithmic}
\end{algorithm}
\section{Pseudo-code for seq-CURL}\label{pseudo-code-seqcurl}
\begin{lstlisting}
def sample_seq_curl(self):

    idxs = np.random.randint(
        0, self.capacity if self.full else self.idx, size=self.batch_size
    )

    obses = self.obses[idxs]
    next_obses = self.next_obses[idxs]

    obses = random_crop(obses, self.image_size)
    next_obses = random_crop(next_obses, self.image_size)

    obses = torch.as_tensor(obses, device=self.device).float()
    next_obses = torch.as_tensor(
        next_obses, device=self.device
    ).float()
    actions = torch.as_tensor(self.actions[idxs], device=self.device)
    rewards = torch.as_tensor(self.rewards[idxs], device=self.device)
    not_dones = torch.as_tensor(self.not_dones[idxs], device=self.device)

    bsz = 8
    length = 32

    idxs = np.random.randint(
        0, self.capacity - length if self.full else self.idx - length,
        size=bsz
    )

    idxs = idxs.reshape(-1, 1)
    step = np.arange(length).reshape(1, -1)
    idxs = idxs + step
    obs_x = self.obses[idxs]
    obs_z = obs_x.copy()
    obs_x = obs_x.reshape(-1, *obs_x.shape[2:])
    obs_z = obs_z.reshape(-1, *obs_z.shape[2:])
    obs_x = random_crop(obs_x, self.image_size)
    obs_z = random_crop(obs_z, self.image_size)
    obs_x = torch.as_tensor(obs_x, device=self.device).float()
    obs_z = torch.as_tensor(obs_z, device=self.device).float()

    cpc_kwargs = dict(obs_anchor=obs_x, obs_pos=obs_z,
                        time_anchor=None, time_pos=None)

    return obses, actions, rewards, next_obses, not_dones, cpc_kwargs
\end{lstlisting}

\end{document}

%% file: math_commands.tex
\usepackage{amsmath,amsfonts,bm}

\def\eqref#1{equation~\ref{#1}}

\def\1{\bm{1}}

\DeclareMathAlphabet{\mathsfit}{\encodingdefault}{\sfdefault}{m}{sl}
\SetMathAlphabet{\mathsfit}{bold}{\encodingdefault}{\sfdefault}{bx}{n}













%% file: iclr2021_conference.bbl
\begin{thebibliography}{39}
\providecommand{\natexlab}[1]{#1}
\providecommand{\url}[1]{\texttt{#1}}
\expandafter\ifx\csname urlstyle\endcsname\relax
  \providecommand{\doi}[1]{doi: #1}\else
  \providecommand{\doi}{doi: \begingroup \urlstyle{rm}\Url}\fi

\bibitem[Ba et~al.(2016)Ba, Kiros, and Hinton]{ba2016layer}
Jimmy~Lei Ba, Jamie~Ryan Kiros, and Geoffrey~E Hinton.
\newblock Layer normalization.
\newblock \emph{arXiv preprint arXiv:1607.06450}, 2016.

\bibitem[Bellemare et~al.(2013)Bellemare, Naddaf, Veness, and
  Bowling]{bellemare2013arcade}
Marc~G Bellemare, Yavar Naddaf, Joel Veness, and Michael Bowling.
\newblock The arcade learning environment: An evaluation platform for general
  agents.
\newblock \emph{Journal of Artificial Intelligence Research}, 47:\penalty0
  253--279, 2013.

\bibitem[Bousmalis et~al.(2018)Bousmalis, Irpan, Wohlhart, Bai, Kelcey,
  Kalakrishnan, Downs, Ibarz, Pastor, Konolige, et~al.]{bousmalis2018using}
Konstantinos Bousmalis, Alex Irpan, Paul Wohlhart, Yunfei Bai, Matthew Kelcey,
  Mrinal Kalakrishnan, Laura Downs, Julian Ibarz, Peter Pastor, Kurt Konolige,
  et~al.
\newblock Using simulation and domain adaptation to improve efficiency of deep
  robotic grasping.
\newblock In \emph{2018 IEEE international conference on robotics and
  automation (ICRA)}, pp.\  4243--4250. IEEE, 2018.

\bibitem[Devlin et~al.(2018)Devlin, Chang, Lee, and Toutanova]{devlin2018bert}
Jacob Devlin, Ming-Wei Chang, Kenton Lee, and Kristina Toutanova.
\newblock Bert: Pre-training of deep bidirectional transformers for language
  understanding.
\newblock \emph{arXiv preprint arXiv:1810.04805}, 2018.

\bibitem[Fortunato et~al.(2017)Fortunato, Azar, Piot, Menick, Osband, Graves,
  Mnih, Munos, Hassabis, Pietquin, et~al.]{fortunato2017noisy}
Meire Fortunato, Mohammad~Gheshlaghi Azar, Bilal Piot, Jacob Menick, Ian
  Osband, Alex Graves, Vlad Mnih, Remi Munos, Demis Hassabis, Olivier Pietquin,
  et~al.
\newblock Noisy networks for exploration.
\newblock \emph{arXiv preprint arXiv:1706.10295}, 2017.

\bibitem[Gu et~al.(2017)Gu, Holly, Lillicrap, and Levine]{gu2017deep}
Shixiang Gu, Ethan Holly, Timothy Lillicrap, and Sergey Levine.
\newblock Deep reinforcement learning for robotic manipulation with
  asynchronous off-policy updates.
\newblock In \emph{2017 IEEE international conference on robotics and
  automation (ICRA)}, pp.\  3389--3396. IEEE, 2017.

\bibitem[Haarnoja et~al.(2018)Haarnoja, Zhou, Abbeel, and
  Levine]{haarnoja2018soft}
Tuomas Haarnoja, Aurick Zhou, Pieter Abbeel, and Sergey Levine.
\newblock Soft actor-critic: Off-policy maximum entropy deep reinforcement
  learning with a stochastic actor.
\newblock \emph{arXiv preprint arXiv:1801.01290}, 2018.

\bibitem[Hafner et~al.(2019{\natexlab{a}})Hafner, Lillicrap, Ba, and
  Norouzi]{hafner2019dream}
Danijar Hafner, Timothy Lillicrap, Jimmy Ba, and Mohammad Norouzi.
\newblock Dream to control: Learning behaviors by latent imagination.
\newblock \emph{arXiv preprint arXiv:1912.01603}, 2019{\natexlab{a}}.

\bibitem[Hafner et~al.(2019{\natexlab{b}})Hafner, Lillicrap, Fischer, Villegas,
  Ha, Lee, and Davidson]{hafner2019learning}
Danijar Hafner, Timothy Lillicrap, Ian Fischer, Ruben Villegas, David Ha,
  Honglak Lee, and James Davidson.
\newblock Learning latent dynamics for planning from pixels.
\newblock In \emph{International Conference on Machine Learning}, pp.\
  2555--2565, 2019{\natexlab{b}}.

\bibitem[Hassani \& Khasahmadi(2020)Hassani and
  Khasahmadi]{hassani2020contrastive}
Kaveh Hassani and Amir~Hosein Khasahmadi.
\newblock Contrastive multi-view representation learning on graphs.
\newblock \emph{arXiv preprint arXiv:2006.05582}, 2020.

\bibitem[He et~al.(2020)He, Fan, Wu, Xie, and Girshick]{he2020momentum}
Kaiming He, Haoqi Fan, Yuxin Wu, Saining Xie, and Ross Girshick.
\newblock Momentum contrast for unsupervised visual representation learning.
\newblock In \emph{Proceedings of the IEEE/CVF Conference on Computer Vision
  and Pattern Recognition}, pp.\  9729--9738, 2020.

\bibitem[H{\'e}naff et~al.(2019)H{\'e}naff, Srinivas, De~Fauw, Razavi, Doersch,
  Eslami, and Oord]{henaff2019data}
Olivier~J H{\'e}naff, Aravind Srinivas, Jeffrey De~Fauw, Ali Razavi, Carl
  Doersch, SM~Eslami, and Aaron van~den Oord.
\newblock Data-efficient image recognition with contrastive predictive coding.
\newblock \emph{arXiv preprint arXiv:1905.09272}, 2019.

\bibitem[Hessel et~al.(2018)Hessel, Modayil, Van~Hasselt, Schaul, Ostrovski,
  Dabney, Horgan, Piot, Azar, and Silver]{hessel2018rainbow}
Matteo Hessel, Joseph Modayil, Hado Van~Hasselt, Tom Schaul, Georg Ostrovski,
  Will Dabney, Dan Horgan, Bilal Piot, Mohammad Azar, and David Silver.
\newblock Rainbow: Combining improvements in deep reinforcement learning.
\newblock In \emph{Thirty-Second AAAI Conference on Artificial Intelligence},
  2018.

\bibitem[Hochreiter \& Schmidhuber(1997)Hochreiter and
  Schmidhuber]{hochreiter1997long}
Sepp Hochreiter and J{\"u}rgen Schmidhuber.
\newblock Long short-term memory.
\newblock \emph{Neural computation}, 9\penalty0 (8):\penalty0 1735--1780, 1997.

\bibitem[Jaderberg et~al.(2016)Jaderberg, Mnih, Czarnecki, Schaul, Leibo,
  Silver, and Kavukcuoglu]{jaderberg2016reinforcement}
Max Jaderberg, Volodymyr Mnih, Wojciech~Marian Czarnecki, Tom Schaul, Joel~Z
  Leibo, David Silver, and Koray Kavukcuoglu.
\newblock Reinforcement learning with unsupervised auxiliary tasks.
\newblock \emph{arXiv preprint arXiv:1611.05397}, 2016.

\bibitem[Kaiser et~al.(2019)Kaiser, Babaeizadeh, Milos, Osinski, Campbell,
  Czechowski, Erhan, Finn, Kozakowski, Levine, et~al.]{kaiser2019model}
Lukasz Kaiser, Mohammad Babaeizadeh, Piotr Milos, Blazej Osinski, Roy~H
  Campbell, Konrad Czechowski, Dumitru Erhan, Chelsea Finn, Piotr Kozakowski,
  Sergey Levine, et~al.
\newblock Model-based reinforcement learning for atari.
\newblock \emph{arXiv preprint arXiv:1903.00374}, 2019.

\bibitem[Khurana et~al.(2020)Khurana, Laurent, and Glass]{khurana2020cstnet}
Sameer Khurana, Antoine Laurent, and James Glass.
\newblock Cstnet: Contrastive speech translation network for self-supervised
  speech representation learning.
\newblock \emph{arXiv preprint arXiv:2006.02814}, 2020.

\bibitem[Kielak(2020)]{kielak2020recent}
Kacper Kielak.
\newblock Do recent advancements in model-based deep reinforcement learning
  really improve data efficiency?
\newblock \emph{arXiv preprint arXiv:2003.10181}, 2020.

\bibitem[Kingma \& Ba(2014)Kingma and Ba]{kingma2014adam}
Diederik~P Kingma and Jimmy Ba.
\newblock Adam: A method for stochastic optimization.
\newblock \emph{arXiv preprint arXiv:1412.6980}, 2014.

\bibitem[Kostrikov et~al.(2020)Kostrikov, Yarats, and
  Fergus]{kostrikov2020image}
Ilya Kostrikov, Denis Yarats, and Rob Fergus.
\newblock Image augmentation is all you need: Regularizing deep reinforcement
  learning from pixels.
\newblock \emph{arXiv preprint arXiv:2004.13649}, 2020.

\bibitem[Lample \& Conneau(2019)Lample and Conneau]{lample2019cross}
Guillaume Lample and Alexis Conneau.
\newblock Cross-lingual language model pretraining.
\newblock \emph{arXiv preprint arXiv:1901.07291}, 2019.

\bibitem[Laskin et~al.(2020)Laskin, Lee, Stooke, Pinto, Abbeel, and
  Srinivas]{laskin2020reinforcement}
Michael Laskin, Kimin Lee, Adam Stooke, Lerrel Pinto, Pieter Abbeel, and
  Aravind Srinivas.
\newblock Reinforcement learning with augmented data.
\newblock \emph{arXiv preprint arXiv:2004.14990}, 2020.

\bibitem[Lee et~al.(2019)Lee, Nagabandi, Abbeel, and Levine]{lee2019stochastic}
Alex~X Lee, Anusha Nagabandi, Pieter Abbeel, and Sergey Levine.
\newblock Stochastic latent actor-critic: Deep reinforcement learning with a
  latent variable model.
\newblock \emph{arXiv preprint arXiv:1907.00953}, 2019.

\bibitem[Li et~al.(2019)Li, Zhang, Bian, Tong, and Liu]{li2019cooperative}
Xihan Li, Jia Zhang, Jiang Bian, Yunhai Tong, and Tie-Yan Liu.
\newblock A cooperative multi-agent reinforcement learning framework for
  resource balancing in complex logistics network.
\newblock In \emph{Proceedings of the 18th International Conference on
  Autonomous Agents and MultiAgent Systems}, pp.\  980--988, 2019.

\bibitem[Mnih et~al.(2013)Mnih, Kavukcuoglu, Silver, Graves, Antonoglou,
  Wierstra, and Riedmiller]{mnih2013playing}
Volodymyr Mnih, Koray Kavukcuoglu, David Silver, Alex Graves, Ioannis
  Antonoglou, Daan Wierstra, and Martin Riedmiller.
\newblock Playing atari with deep reinforcement learning.
\newblock \emph{arXiv preprint arXiv:1312.5602}, 2013.

\bibitem[Mnih et~al.(2016)Mnih, Badia, Mirza, Graves, Lillicrap, Harley,
  Silver, and Kavukcuoglu]{mnih2016asynchronous}
Volodymyr Mnih, Adria~Puigdomenech Badia, Mehdi Mirza, Alex Graves, Timothy
  Lillicrap, Tim Harley, David Silver, and Koray Kavukcuoglu.
\newblock Asynchronous methods for deep reinforcement learning.
\newblock In \emph{International conference on machine learning}, pp.\
  1928--1937, 2016.

\bibitem[Oord et~al.(2018)Oord, Li, and Vinyals]{oord2018representation}
Aaron van~den Oord, Yazhe Li, and Oriol Vinyals.
\newblock Representation learning with contrastive predictive coding.
\newblock \emph{arXiv preprint arXiv:1807.03748}, 2018.

\bibitem[Shelhamer et~al.(2016)Shelhamer, Mahmoudieh, Argus, and
  Darrell]{shelhamer2016loss}
Evan Shelhamer, Parsa Mahmoudieh, Max Argus, and Trevor Darrell.
\newblock Loss is its own reward: Self-supervision for reinforcement learning.
\newblock \emph{arXiv preprint arXiv:1612.07307}, 2016.

\bibitem[Silver et~al.(2018)Silver, Hubert, Schrittwieser, Antonoglou, Lai,
  Guez, Lanctot, Sifre, Kumaran, Graepel, et~al.]{silver2018general}
David Silver, Thomas Hubert, Julian Schrittwieser, Ioannis Antonoglou, Matthew
  Lai, Arthur Guez, Marc Lanctot, Laurent Sifre, Dharshan Kumaran, Thore
  Graepel, et~al.
\newblock A general reinforcement learning algorithm that masters chess, shogi,
  and go through self-play.
\newblock \emph{Science}, 362\penalty0 (6419):\penalty0 1140--1144, 2018.

\bibitem[Srinivas et~al.(2020)Srinivas, Laskin, and Abbeel]{srinivas2020curl}
Aravind Srinivas, Michael Laskin, and Pieter Abbeel.
\newblock Curl: Contrastive unsupervised representations for reinforcement
  learning.
\newblock \emph{arXiv preprint arXiv:2004.04136}, 2020.

\bibitem[Sun et~al.(2019)Sun, Hoffmann, Verma, and Tang]{sun2019infograph}
Fan-Yun Sun, Jordan Hoffmann, Vikas Verma, and Jian Tang.
\newblock Infograph: Unsupervised and semi-supervised graph-level
  representation learning via mutual information maximization.
\newblock \emph{arXiv preprint arXiv:1908.01000}, 2019.

\bibitem[Tassa et~al.(2020)Tassa, Tunyasuvunakool, Muldal, Doron, Liu, Bohez,
  Merel, Erez, Lillicrap, and Heess]{tassa2020dmcontrol}
Yuval Tassa, Saran Tunyasuvunakool, Alistair Muldal, Yotam Doron, Siqi Liu,
  Steven Bohez, Josh Merel, Tom Erez, Timothy Lillicrap, and Nicolas Heess.
\newblock dm\_control: Software and tasks for continuous control, 2020.

\bibitem[Tian et~al.(2019)Tian, Krishnan, and Isola]{tian2019contrastive}
Yonglong Tian, Dilip Krishnan, and Phillip Isola.
\newblock Contrastive multiview coding.
\newblock \emph{arXiv preprint arXiv:1906.05849}, 2019.

\bibitem[{Todorov} et~al.(2012){Todorov}, {Erez}, and {Tassa}]{6386109}
E.~{Todorov}, T.~{Erez}, and Y.~{Tassa}.
\newblock Mujoco: A physics engine for model-based control.
\newblock In \emph{2012 IEEE/RSJ International Conference on Intelligent Robots
  and Systems}, pp.\  5026--5033, 2012.

\bibitem[van Hasselt et~al.(2019)van Hasselt, Hessel, and
  Aslanides]{van2019use}
Hado~P van Hasselt, Matteo Hessel, and John Aslanides.
\newblock When to use parametric models in reinforcement learning?
\newblock In \emph{Advances in Neural Information Processing Systems}, pp.\
  14322--14333, 2019.

\bibitem[Vaswani et~al.(2017)Vaswani, Shazeer, Parmar, Uszkoreit, Jones, Gomez,
  Kaiser, and Polosukhin]{vaswani2017attention}
Ashish Vaswani, Noam Shazeer, Niki Parmar, Jakob Uszkoreit, Llion Jones,
  Aidan~N Gomez, {\L}ukasz Kaiser, and Illia Polosukhin.
\newblock Attention is all you need.
\newblock In \emph{Advances in neural information processing systems}, pp.\
  5998--6008, 2017.

\bibitem[Vinyals et~al.(2019)Vinyals, Babuschkin, Czarnecki, Mathieu, Dudzik,
  Chung, Choi, Powell, Ewalds, Georgiev, et~al.]{vinyals2019grandmaster}
Oriol Vinyals, Igor Babuschkin, Wojciech~M Czarnecki, Micha{\"e}l Mathieu,
  Andrew Dudzik, Junyoung Chung, David~H Choi, Richard Powell, Timo Ewalds,
  Petko Georgiev, et~al.
\newblock Grandmaster level in starcraft ii using multi-agent reinforcement
  learning.
\newblock \emph{Nature}, 575\penalty0 (7782):\penalty0 350--354, 2019.

\bibitem[Wu et~al.(2018)Wu, Xiong, Yu, and Lin]{wu2018unsupervised}
Zhirong Wu, Yuanjun Xiong, Stella Yu, and Dahua Lin.
\newblock Unsupervised feature learning via non-parametric instance-level
  discrimination.
\newblock \emph{arXiv preprint arXiv:1805.01978}, 2018.

\bibitem[Yarats et~al.(2019)Yarats, Zhang, Kostrikov, Amos, Pineau, and
  Fergus]{yarats2019improving}
Denis Yarats, Amy Zhang, Ilya Kostrikov, Brandon Amos, Joelle Pineau, and Rob
  Fergus.
\newblock Improving sample efficiency in model-free reinforcement learning from
  images.
\newblock \emph{arXiv preprint arXiv:1910.01741}, 2019.

\end{thebibliography}
